\documentclass{article}

\usepackage[final, nonatbib]{neurips_2023}

\usepackage[utf8]{inputenc} % allow utf-8 input
\usepackage[T1]{fontenc}    % use 8-bit T1 fonts
\usepackage{hyperref}       % hyperlinks
\usepackage{url}            % simple URL typesetting
\usepackage{booktabs}       % professional-quality tables
\usepackage{amsfonts}       % blackboard math symbols
\usepackage{nicefrac}       % compact symbols for 1/2, etc.
\usepackage{microtype}      % microtypography
\usepackage{xcolor}         % colors 

\usepackage{subcaption}
\usepackage{graphicx} 
\usepackage{amsmath}
\usepackage{amssymb}
\usepackage{mathtools}
\usepackage{amsthm}
\usepackage{multirow}
\usepackage{enumitem}

\setlist{leftmargin=7mm}

\DeclareMathOperator*{\argmax}{arg\,max}

\title{NEO-KD: Knowledge-Distillation-Based Adversarial Training for Robust Multi-Exit Neural Networks}

\author{%
Seokil Ham$^{1}$ \quad Jungwuk Park$^{1}$ \quad Dong-Jun Han$^2$\thanks{Corresponding author.} \quad Jaekyun Moon$^1$  \\
$^1$KAIST \quad $^2$Purdue University\\
\texttt{\{gkatjrdlf, savertm\}@kaist.ac.kr}, \ \ \texttt{han762@purdue.edu}, \ \ \texttt{jmoon@kaist.edu}\\
}

\begin{document}

\maketitle

\begin{abstract}
  While multi-exit neural networks are regarded as a promising solution for making efficient inference via early exits, combating adversarial attacks remains a challenging problem. In multi-exit networks, due to the high dependency among different submodels, an adversarial example targeting a specific exit not only degrades the performance of the target exit but also reduces the performance of all other exits concurrently. This makes multi-exit networks highly vulnerable to simple adversarial attacks.   %In this paper, we propose NEO-KD, a knowledge-distillation-based adversarial training strategy that tackles this fundamental challenge of multi-exit  networks with two key contributions. 
    In this paper, we propose NEO-KD, a knowledge-distillation-based
adversarial training strategy that tackles this fundamental challenge based on two key contributions. NEO-KD first resorts to neighbor knowledge distillation to  guide the output of the adversarial examples to tend to the ensemble outputs of neighbor exits of clean data. NEO-KD also employs exit-wise orthogonal knowledge distillation  for reducing adversarial transferability across different submodels. The result is a significantly improved robustness against adversarial attacks.  
  Experimental results on various datasets/models show that our method  achieves the best adversarial accuracy with reduced computation budgets, compared to the baselines relying on existing adversarial training or knowledge distillation techniques for multi-exit networks.
\end{abstract}

\section{Introduction}

Multi-exit neural networks are receiving significant attention  \cite{han2023improving,huang2017multi,teerapittayanon2016branchynet,wang2021harmonized,xing2020early,zhou2020bert} for their ability to make dynamic predictions in resource-constrained applications. Instead of making predictions at the final output of the full model, a faster prediction can be made at an earlier exit depending on the current time budget or computing budget.  In this sense, a multi-exit network can be viewed as an architecture having multiple submodels, where each submodel consists of parameters from the input of the   model to the output of a specific exit. These submodels are highly correlated as they share some model parameters.  
It is also well-known that the performance of all submodels can be improved
by distilling the knowledge of the last exit to other exits, i.e., via self-distillation \cite{kouris2022multi,li2019improved,phuong2019distillation,wang2021harmonized}.
There have also been efforts to address the adversarial attack issues in the context of multi-exit networks \cite{chen2022towards,hu2019triple}.
%It is also well-known that knowledge distillation (or self-distillation) strategies can further improve the prediction performance of exits by distilling the prediction of the last exit to others.

Providing robustness against adversarial attacks is especially   challenging  in multi-exit networks: since different submodels have high correlations by sharing parameters, an adversarial example
targeting a specific exit can significantly degrade the performance of other submodels. In other words, an adversarial example can have strong \textit{adversarial transferability} across different submodels, making the model   highly vulnerable to  simple adversarial attacks (e.g., an adversarial attack targeting a single exit).

\textbf{Motivation.} Only a few prior works have focused on adversarial defense strategies for multi-exit networks \cite{chen2022towards,hu2019triple}. The authors of  \cite{hu2019triple}  focused on generating adversarial examples tailored to multi-exit networks (e.g., generate samples via max-average attack), and trained the model to minimize the sum of clean and adversarial losses of all exits. Given the adversarial example constructed in  \cite{hu2019triple}, the authors of  \cite{chen2022towards} proposed a regularization term to reduce the weights of  the classifier at each exit  during training.  However, existing adversarial defense strategies \cite{chen2022towards,hu2019triple} do not  directly handle the high correlations among different submodels, resulting in high
adversarial transferability and limited robustness in multi-exit networks. To tackle this difficulty, we take a knowledge-distillation-based approach in a fashion orthogonal to prior works
 \cite{chen2022towards,hu2019triple}. Some previous studies \cite{goldblum2020adversarially,papernot2016distillation,zhu2021reliable,zi2021revisiting} have shown that knowledge distillation can be utilized for improving the robustness of the model in conventional single-exit networks. However, although there are extensive existing works on self-distillation for training multi-exit networks using clean data \cite{kouris2022multi,li2019improved,phuong2019distillation,wang2021harmonized}, it is currently unknown how distillation techniques should be utilized for adversarial training of multi-exit networks. Moreover, when the existing distillation-based schemes are applied to multi-exit networks, the dependencies among submodels become higher since the same output (e.g., the knowledge of the last exit) is distilled to all sub-models. Motivated by these limitations, we pose the following questions:  \textit{How can we take advantage of  knowledge-distillation to improve adversarial   robustness of multi-exit networks? At the same time, how can we reduce  adversarial transferability across different submodels in multi-exit networks?}
 
%not take advantage of knowledge distillation, which is a de facto standard in multi-exit neural networks;  knowledge distillation techniques to improve the adversarial accuracy has not been well explored in previous works. Moreover, existing works do not directly handle the high correlations among different submodels, which results in high adversarial transferability.  These two issues limit the robustness of multi-exit neural networks against adversarial attacks.

\textbf{Main contributions.} To handle these questions, we propose NEO-KD, a knowledge-distillation-based adversarial training strategy highly tailored to
robust multi-exit neural networks. Our solution is two-pronged: neighbor
knowledge distillation and exit-wise orthogonal knowledge distillation. 

 \begin{itemize}
  \item Given a specific exit, the first part of our solution, neighbor knowledge distillation (NKD),  distills the ensembled prediction of neighbor exits of clean data to the prediction of the adversarial example at the corresponding exit, as shown in Figure \ref{figure: NKD}.
%distills the prediction of  the original data (or clean data) at the last exit to the predictions of  both adversarial/clean samples  at all exits.  
This method guides the output of  adversarial examples to follow the outputs of clean data, improving robustness against adversarial attacks. By ensembling the neighbor predictions of clean data before distillation, NKD provides higher quality features to the corresponding exits compared to the scheme distilling with only one exit in the same position.  
%Since NKD provides not only the features of  the exit in the same position but also the features of neighbor exits, each exit is distilled with higher quality feature  compared to the scheme distilling with only one exit in the same position.  
%Moreover, by distilling different teacher predictions to each exit, NKD reduces the adversarial transferability across the submodels in the multi-exit network, which has not been addressed in prior works \cite{chen2022towards,hu2019triple}.
%In addition, NKD considers adversarial transferability to become low through distilling different teacher predictions to each exit, which is not addressed at prior works \cite{hu2019triple,chen2022towards}.
\item The  second focus of our solution,  exit-wise  orthogonal knowledge distillation (EOKD), mainly  aims at reducing adversarial transferability across different submodels.  This part is another unique contribution of our work compared to existing methods on robust multi-exit networks  \cite{chen2022towards,hu2019triple} (that suffer from high adversarial transferability) or self-distillation-based multi-exit networks \cite{kouris2022multi,li2019improved,phuong2019distillation,wang2021harmonized}  (that further increase  adversarial transferability). % with two characteristics: exit-wise distillation and orthogonal distillation. 
In our EOKD, the output of   clean data at  the $i$-th exit is distilled to the  output of the adversarial sample at  the $i$-th exit, in an exit-wise manner. During this exit-wise distillation process, we encourage the non-ground-truth predictions of individual exits to be mutually orthogonal, by providing orthogonal soft labels to each exit  as described in   Figure \ref{figure: EOKD}. By weakening the dependencies
among different exit outputs,  EOKD reduces the  adversarial  transferability across all submodels in the network, which leads to an improved robustness against adversarial attacks.  
 \end{itemize}
The NKD and EOKD components of our architectural solution work together to reduce adversarial transferability across different submodels in the network while correctly guiding the predictions of the adversarial examples at each exit. Experimental results on various datasets show that the proposed strategy achieves the best adversarial accuracy with reduced computation budgets, compared to existing adversarial training methods for multi-exit networks. Our solution  is a plug-and-play method, which can be used in conjunction with existing   training strategies tailored to multi-exit networks. 
 % \vspace{-1mm}

\section{Related Works}
  %\vspace{-1mm}

%\subsection{Knowledge Distillation for Training  Multi-Exit   Networks}
\textbf{Knowledge distillation for multi-exit networks.} 
Multi-exit neural networks \cite{han2023improving,huang2017multi,teerapittayanon2016branchynet,wang2021harmonized,xing2020early,zhou2020bert} aim at making efficient inference via early exits in resource-constrained applications. %is a deep neural network which contains several early exits for various purposes . %The additional early exits make fast inference in time-limited situation and budget-limited situation. MSDNet \cite{huang2017multi} adopts dense connectivities to extract features for later classifiers as well as early classifiers at early stage blocks, and multi-scale features to utilize coarse and fine features at early classifiers. Since MSDNet considers the shortages of multi-exit neural network and achieves higher performance than other prior works, we adopt MSDNet as our baseline architecture.
In the multi-exit
network literature,  it is well-known that distilling the knowledge of  the last exit to others significantly improves the overall performance on clean data without an external teacher network, i.e., via self-distillation \cite{kouris2022multi,li2019improved,phuong2019distillation,wang2021harmonized}. However,    %self-distillation strategies to  improve  adversarial robustness of multi-exit networks   have not been studied in the literature;
it is currently unclear  how adversarial training  can benefit from self-distillation in multi-exit networks. One challenge is that simply applying existing self-distillation techniques increases adversarial transferability across different submodels, since the same knowledge from the last exit is distilled to all  other exits, increasing  dependency among different submodels in the network. Compared to the existing ideas, our contribution is to develop
a self-distillation strategy that does not increase the dependency of submodels as much; this helps reduce adversarial transferability of the multi-exit network for better robustness.

\textbf{Improving adversarial robustness.}
%\textcolor{blue}{To improve adversarial robustness in traditional single-exit networks, most existing defense methods \cite{zhang2019theoretically, dong2021exploring, dong2023enemy} have mainly focused on creating new adversarial training losses restricted to single-exit network. Several other works have utilized the concept of knowledge distillation  \cite{goldblum2020adversarially,papernot2016distillation,zhu2021reliable,zi2021revisiting} showing that distilling the knowledge of the teacher network can improve robustness of the student network.  Especially in \cite{goldblum2020adversarially}, given a robust teacher network, it is shown that robustness of the teacher network can be distilled to the student network during adversarial training. However, as these works only consider single-exit networks, adversarial transferability between exits has not been a key issue in such works.} 
Most existing defense methods \cite{dong2023enemy,dong2021exploring,zhang2019theoretically} have mainly focused on creating new adversarial training losses  tailored  to single-exit  networks.  Several other works have utilized the concept of knowledge distillation  \cite{goldblum2020adversarially,papernot2016distillation,zhu2021reliable,zi2021revisiting} showing that distilling the knowledge of the teacher network can improve robustness of the student network. Especially in \cite{goldblum2020adversarially}, given a  teacher network, it is shown that robustness of the teacher network can be distilled to the student network during adversarial training. Compared to these works, our approach  can be viewed as a new self-distillation strategy for multi-exit networks where teacher/student models are trained together. More importantly, adversarial transferability across different submodels has not been an
issue in previous works as the focus there has been on the single-exit network.  In contrast, in our multi-exit setup, all submodels sharing some model parameters require extra robustness against adversarial attacks; this motivates us to propose exit-wise orthogonal knowledge distillation,
to reduce adversarial transferability among different submodels.
%} %More importantly, different from previous works constructing a single robust student network, in our setup, all submodels  in the multi-exit network (sharing some parameters) should gain robust against adversarial attacks, and thus the adversarial transferability should be taken into account.  }

Some prior works \cite{lee2022graddiv,pang2019improving,yang2020dverge,yang2021trs} aim at improving  adversarial robustness of the ensemble model, by reducing adversarial transferability  across individual models. 
%Adversarial transferability means how adversarial attack generated from a model is fatal to the other models.
%Since several submodels in an ensemble model are likely to have high correlations, an ensemble model is sensitive to adversarial transferability.
%Then, if an ensemble model has low adversarial transferability, submodels in the ensemble model will output independent predictions, which result in a robust final prediction of the ensemble model.
Specifically, the adaptive diversity-promoting regularizer proposed in \cite{pang2019improving} regularizes the non-maximal predictions of individual models
to be mutually orthogonal, and the maximal term is used to compute the loss as usual.
%This regularization promotes the diversity of the ensemble model and reduces adversarial transferability, which leads better robustness for the ensemble model.
While the previous work focuses on reducing the transferability among different models having independent parameters, in a multi-exit network setup, the problem becomes more challenging in that all submodels have  some  shared parameters, making the models to be highly correlated. To handle this issue, we specifically take advantage of knowledge distillation in an exit-wise manner, which can further reduce the dependency among different submodels in the multi-exit network.%To handle this issue, we propose a new exit-wise distillation method which can effectively reduce the dependency among  different submodels  in the multi-exit network. 
%\textcolor{red}{While we also adopt the concept of mutual orthogonality in our EOKD,   different from this work, we propose a knowledge distillation based approach tailored to multi-exit network for mutual orthogonality across exits, not the ensemble models.}

\textbf{Adversarial training for multi-exit  networks.}
%\subsection{Adversarial Training in Multi-Exit  Networks} 
When focused on  multi-exit networks, only a few prior works considered the adversarial attack issue in the literature \cite{chen2022towards,haque2020ilfo,hong2020panda,hu2019triple}.
The authors of \cite{haque2020ilfo,hong2020panda} focused on generating slowdown  attacks in multi-exit networks rather than defense strategies.
%Then, these prior works are not aligned with our purpose which makes multi-exit network robust against general adversarial attack.
%The authors of \cite{haque2020ilfo, hong2020panda} proposed various attacks that can reduce the performance of multi-exit neural networks. 
%\cite{hu2019triple} proposed an adversarial training method for multi-exit network by considering both original data and adversarial examples simultaneously, where the adversarial examples are generated targeting a specific exit (single attack) or multiple exits (average attack and max-average attack).
In \cite{hu2019triple},   the authors proposed an adversarial training  strategy  by generating adversarial examples targeting a specific exit (single attack) or multiple exits (average attack and max-average attack). However, (i) \cite{hu2019triple} does not take advantage of knowledge distillation during training and (ii) \cite{hu2019triple} does not directly handle the high correlations among different submodels, which can result in high adversarial transferability. Our solution overcomes these limitations by reducing adversarial transferability while correctly guiding the predictions of adversarial examples at each exit, via self knowledge distillation. %To handle these limitations, we develop a knowledge distillation based   strategy to reduce  the adversarial transferability and improve the  adversarial accuracy of the network. %The adversarial training method utilizing the average attack and max-average attack significantly improves the robustness of multi-exit neural network. 
%Our work  is orthogonal to this approach, in that our work additionally utilizes knowledge distillation for adversarial training to make multi-exit network more robust.  
%Built upon the work   of \cite{hu2019triple}, our knowledge distillation strategies further improve adversarial test accuracy of multi-exit network. 

%i) do not take advantage of knowledge distillation during training and (ii) do not directly handle the high correlations among different submodels, which results in high adversarial transferability. 

\vspace{-1.5mm}
\section{Proposed NEO-KD Algorithm}
\vspace{-1.5mm}
Consider a multi-exit   network with $L$ exits, which is composed of $L$ blocks $\{\phi_i\}_{i=1}^L$ and $L$ classifiers $\{w_i\}_{i=1}^L$. Given the input data $x$, the output of the $i$-th exit is denoted as $f_{\theta_i}(x)$, which is parameterized by the $i$-th submodel   $\theta_i = [\phi_1,\dots,\phi_i,w_i]$ that consists  of $i$ blocks and one classifier.  Note that all $L$ submodels produce different predictions [$f_{\theta_{1}}(x),\dots, f_{\theta_{L}}(x)$]. Here, since each submodel shares several blocks with other submodels,   the predictions of any two submodels are highly correlated. % \textcolor{red}{OK until here!}

%Based on these notations,  given the input data $x$, a multi-exit network can produce $L$ different predictions [$f_{\theta_{1}}(x),\dots, f_{\theta_{L}}(x)$].

%where an auxiliary classifier $\phi_i$ is utilized to make predictions at the $i$-th exit. 

%\subsubsection{Problem Setup}
%\textcolor{red}{Multi-exit network has $L$ exits to output predictions at the intermediate of the  network.
%We notate that $i$-th submodel, which outputs result through $i$-th exit, as $f_{\theta_{i}}(\cdot)$ which is parameterized by $\theta_{i}$.
%And each submodel is consist of several blocks and a classifier, which is parameterized by $\phi$ and $w$.
%That is, the parameters of a submodel $\theta$ contain both block parameter $\phi$ and classifier parameter $w$.
%Then, the parameters of $i$-th submodel, $\theta_i$, include block parameters from initial block parameters $\phi_1$ to $i$-th block parameter $\phi_i$ and $i$-th classifier parameters $w_i$.  
%Given input data $x$ and ground truth $y$, a multi-exit network which has $L$ predictions [$f_{\theta_{1}}(x),\dots, f_{\theta_{L}}(x)$].}

\vspace{-1mm}

\subsection{Problem Setup: Adversarial Training in  Multi-Exit Networks}  
\vspace{-1mm}
The first step for adversarial training is to generate adversarial examples. Given $L$ different submodels $\{\theta_i\}_{i=1}^L$, clean data $x$, and the corresponding label $y$,  the adversarial example $x^{adv}$ can be generated  based on  single  attack, max-average attack or average attack, following the process of  \cite{hu2019triple}. More specifically, we have
%While both attacks account for multiple exits within a multi-exit network, their methods differ.
%For a multi-exit network with $L$ exits, the max-average attack first generates $L$ adversarial examples, each targeting a specific exit, utilizing an \textit{attacker algorithm} (e.g., FGSM, PGD). Subsequently, the adversarial example that maximizes the sum of losses across all exits is selected.  On the other hand, average attack generates an adversarial example that maximizes the average loss of all exits through the attacker algorithm. The respective formulations of generating max-average and average attack are as follows:
 \begin{equation}\label{eq: single}
 	\small
	x^{adv}_{single, i}=\argmax_{x' \in \{z : \lvert z-x\rvert_\infty\leq\epsilon\}}\lvert \ell(f_{\theta_i}(x'),y)\rvert,
\end{equation}
\begin{equation}\label{eq: max-average}
	 \small
	x^{adv}_{max}=x^{adv}_{i^*},\enspace \text{where} \enspace i^*=\argmax_i \bigg\lvert\frac{1}{L}\sum_{j=1}^{L} \ell(f_{\theta_{j}}(x^{adv}_{single, i}),y)\bigg\rvert,
\end{equation}
\begin{equation}\label{eq: average}
	\small
	x^{adv}_{avg}=\argmax_{x' \in \{z : \lvert z-x\rvert_\infty\leq\epsilon\}}\bigg\lvert \frac{1}{L}\sum_{j=1}^{L}\ell(f_{\theta_j}(x'),y)\bigg\rvert,
\end{equation}
which correspond to the adversarial example generated by single attack targeting exit $i$, max-average attack, and average attack, respectively. $\epsilon$ denotes the perturbation degree.   In the single attack of  (\ref{eq: single}), the adversarial example  $x^{adv}_{i}$ is generated to maximize the cross-entropy loss $\ell(\cdot, \cdot)$  of the target exit utilizing an \textit{attacker algorithm} (e.g., PGD \cite{madry2018towards}). In the max-average attack of (\ref{eq: max-average}), among the adversarial
examples generated by the single attack for all exits $i=1,2,\dots,L$, the  sample  that maximizes the average loss of all exits is selected. Finally, the average attack in  (\ref{eq: average})  directly generates an adversarial sample that maximizes the average loss of all exits. Based on the generated $x^{adv}$, a typical strategy is to update the model considering both the clean and adversarial losses of all exits as follows:
 \begin{equation}\label{eq: adversarial training}\small
\mathcal{L}=\frac{1}{N}\sum_{j=1}^{N}\sum_{i=1}^{L}[\ell(f_{\theta_{i}}(x_j),y_j)+\ell(f_{\theta_{i}}(x^{adv}_j),y_j)],
\end{equation}
where $N$ is the number of samples in the training set and $x_j^{adv}$ is the adversarial example corresponding to clean sample $x_j$ generated by one of the attacks described above. However, the loss in (\ref{eq: adversarial training}) does not directly consider the correlation  among submodels, which  can potentially increase  adversarial transferability of the multi-exit network.

%\textcolor{red}{However, the loss in (\ref{eq: adversarial training}) does not directly address adversarial transferability among submodels, which is a key challenge of multi-exit neworks. In this work, we take advantage of self-distillation, which can provide a great platform to tackle this issue, as each exit can be distilled with different knowledge to reduce the dependencies among exits. }

\begin{figure*}[t]
 \centering
\begin{subfigure}{.495\textwidth} 
\centering
\includegraphics[width=1\linewidth]{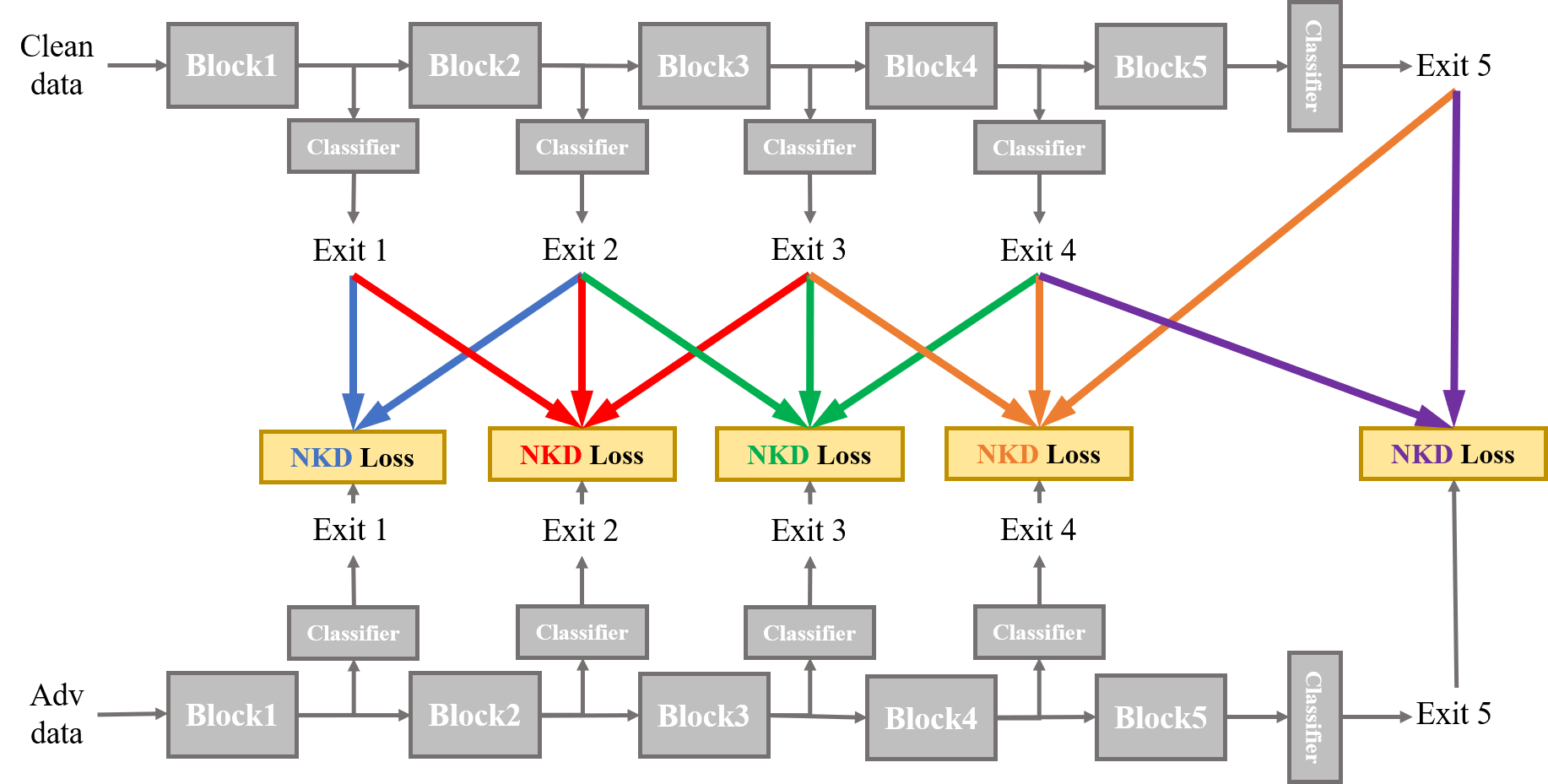}
\caption{Neighbor knowledge distillation}
\label{figure: NKD}
\end{subfigure}
\begin{subfigure}{.495\textwidth}
\centering
\includegraphics[width=1\linewidth]{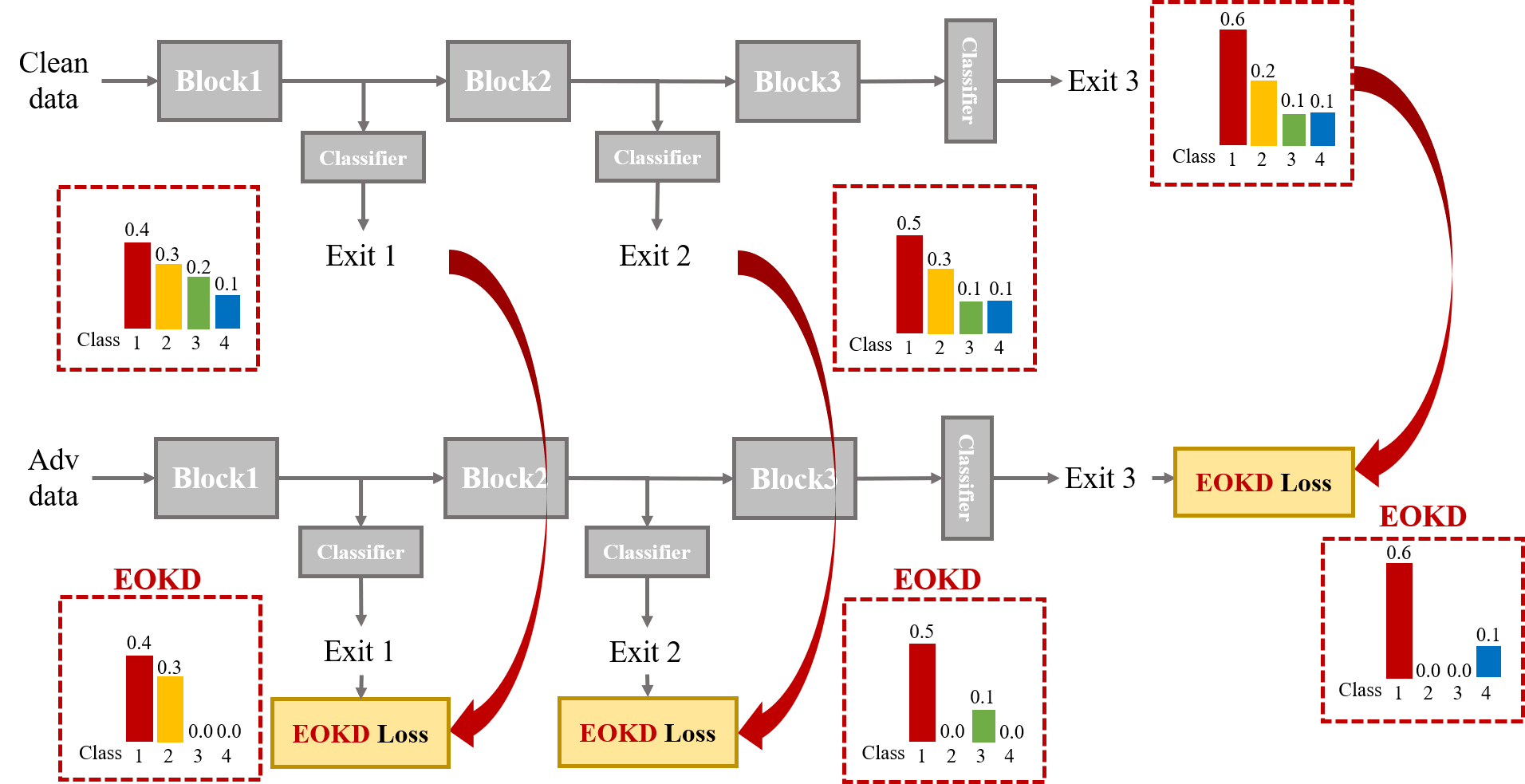}
\caption{Exit-wise orthogonal knowledge distillation}
\label{figure: EOKD}
\end{subfigure}
\vspace{-1mm}
\caption{NEO-KD consists of two parts that together improve the adversarial robustness: NKD and EOKD.
(a) NKD guides the output of the adversarial data to mimic the ensemble outputs of neighbor exits of clean data. (b) EOKD reduces  adversarial transferability of the network by distilling orthogonal knowledge of the clean data to adversarial data for the non-ground-truth predictions, in an exit-wise manner. Although omitted in this figure, EOKD normalizes the likelihood before distilling the soft labels. The overall process operates in a single model, although we consider two cases
depending on the input (clean or adversarial example) for a clear presentation.}
\label{figure: overall method}
\vspace{-3mm}
\end{figure*}

\subsection{Algorithm Description}
%\vspace{-1mm}
%Built upon the loss function in \eqref{eq: adversarial training}, %we propose adversarial exit-wise orthogonal knowledge distillation (AEOKD), 
To tackle the limitations of  prior works \cite{chen2022towards,hu2019triple}, we propose neighbor exit-wise orthogonal knowledge distillation (NEO-KD), a self-distillation strategy tailored to robust multi-exit networks.  To gain insights, we divide our   solution into two distinct components with different roles - neighbor knowledge distillation and exit-wise orthogonal knowledge distillation - and first describe each component separately, and then put together into the overall NEO-KD method.  A  high-level description of our approach is given in  Figure \ref{figure: overall method}. 
\textbf{Neighbor knowledge distillation (NKD).} 
The first component of our solution, NKD,  guides the output feature of adversarial data at each exit to mimic the output feature of   clean data.  Specifically, the proposed   NKD loss of   the $j$-th train sample at  the  $i$-th exit  is written as follows:  
 \begin{equation}
\label{equation: NKD}
NKD_{i,j} = \begin{cases}
\ell(f_{\theta_1}(x^{adv}_j),\frac{1}{2}\sum_{k=1}^{2}f_{\theta_k}(x_j)) &\text{$i=1$}\\
\ell(f_{\theta_L}(x^{adv}_j),\frac{1}{2}\sum_{k=L-1}^{L}f_{\theta_k}(x_j)) &\text{$i=L$}\\
\ell(f_{\theta_i}(x^{adv}_j),\frac{1}{3}\sum_{k=i-1}^{i+1}f_{\theta_k}(x_j)) &\text{otherwise},
\end{cases}
\end{equation}
which can be visualized with the colored arrows as in Figure \ref{figure: NKD}. Different from previous self-knowledge distillation methods, for each exit $i$,
NKD generates a teacher prediction by ensembling (averaging)  the neighbor predictions (i.e., from exit $i-1$ and exit $i+1$) of clean data and distills it to each prediction of adversarial examples.

Compared to other distillation strategies, NKD  takes advantage of only  these \textit{neighbors} during distillation, which has the following key advantages for improving adversarial robustness. First, by ensembling the neighbor predictions of clean data before distillation, NKD provides a higher quality feature of the original data to the corresponding exit; compared to the naive baseline that is distilled with only one exit in the same position (without ensembling), NKD achieves better adversarial accuracy, where the  results are provided in Table \ref{table: number of group}. Secondly, by considering only the neighbors during ensembling, we can distill different teacher predictions to each exit. Different
teacher predictions of NKD also play a role of reducing adversarial transferability compared to the strategies that distill the same prediction (e.g., the last exit) to all exit; ensembling other exits (beyond the   neighbors) increases the dependencies among submodels, resulting in higher adversarial transferability. The corresponding results are  also shown via experiments in Section \ref{section: group number}.

However, a multi-exit network trained with only NKD loss still has a significant room for mitigating adversarial transferability further. In the following, we describe the second part of our solution that solely focuses on reducing  adversarial transferability of multi-exit networks.

\textbf{Exit-wise orthogonal knowledge distillation (EOKD).}  
%To more focus on reducing the adversarial transferability, EOKD is proposed. 
EOKD provides orthogonal soft labels to each exit for the non-ground-truth
predictions, in an exit-wise manner. As can be seen from the red arrows in Figure  \ref{figure: EOKD}, the output of clean data at the $i$-th exit is distilled to the output of adversarial example at the $i$-th exit.  During this exit-wise distillation process, some predictions are discarded to   encourage the non-ground-truth predictions of individual exits to be mutually orthogonal. We randomly allocate the classes of non-ground-truth predictions to each exit for every epoch, which prevents the classifier to be biased compared to the fixed allocation strategy.
%\textcolor{red}{For distilling only orthogonality to each exit, we randomly allocate classes to each exit for not being discarded. Then, EOKD prevents biased classifier, which can be generated when classes are fixed to each exit.} 
The proposed EOKD loss of the $j$-th sample at the $i$-th exit is defined as follows: 
\begin{equation}\label{eq: EOKD}
EOKD_{i,j} = \ell(f_{\theta_i}(x^{adv}_j),O(f_{\theta_i}(x_j))).
\end{equation}
Here, $O(\cdot)$  is  the orthogonal labeling operation to make the non-ground-truth predictions to be  orthogonal across all   exits. For each exit, $O(\cdot)$ randomly selects $\lfloor (C-1)/L \rfloor$ labels among a total of $C$ classes so that   the selected labels  are non-overlapping across different exits (except for the ground-truth label), as in Figure \ref{figure: EOKD}.  Lastly, the probabilities of selected labels are normalized to sum to one. 
 To gain clearer insights, consider a toy example with a 3-exit network (i.e., $L=3$) focusing on a $4$-way classification task (i.e., $C=4$). Let $[p^i_1, p^i_2, p^i_3, p^i_4]$	be the softmax output of the clean sample at the $i$-th exit, for $i=1,2,3$. If class 1 is the ground-truth, the orthogonal labeling operation $O(\cdot)$ jointly produces the following results from each exit: $[\hat{p}^1_1, \hat{p}^1_2, 0, 0]$ from exit 1, $[\hat{p}^2_1, 0, \hat{p}^2_3, 0]$	from exit 2, $[\hat{p}^3_1, 0, 0, \hat{p}^3_4]$	from exit 3, where $\hat{p}$ indicates the normalized probability of $p$ so that the values in each vector sum to one.

 Based on Eq. \eqref{eq: EOKD}, the output of the orthogonal label operation $O(f_{\theta_i}(x_j))$ for the clean data $x_j$ is distilled to $f_{\theta_i}(x^{adv}_j)$ which is the prediction of the adversarial example of the $j$-th sample at the $i$-th exit. This encourages the model to self-distill orthogonally distinct knowledge in an exit-wise manner while keeping the essential knowledge of the ground-truth class. By taking this exit-wise orthogonal distillation approach, EOKD reduces the dependency among different submodels, reducing the adversarial transferability of the network. % and distill to prediction of adversarial examples at each exit.
%\textcolor{red}{
%  This strategy trains each exit to output different likelihood on same data except for the likelihood of answer label.  For example, illustrated in Figure \ref{figure: overall method}, if an input image is an image of class 1, each exit is distilled the prediction of only non-overlapping selected class (class 2 for exit 1, class 3 for exit 2, and class 4 for exit 3) and answer class (class 1). 
%Then, a multi-exit network trained by EOKD results in different likelihood across exits while answer label sustains maximum likelihood.
%Simultaneously, this strategy provides different teacher predictions correspond to each student exit; he $i$-th prediction of clean data distills knowledge to the $i$-th prediction of adversarial examples.
%By adopting exit-wise manner,  each exit can be provided exit-adaptive feature as a teacher prediction for knowledge distillation.
%Finally, these orthogonal manner and exit-wise manner of EOKD mitigate adversarial transferability and improve robustness of multi-exit neural network.}

%Equation \ref{eq: EOKD} describes EOKD objective function when $i$-th exit prediction of $j$-th original image is modified by orthogonal labeling algorithm and distilled to $i$-th exit prediction of $j$-th adversarial example.

%\subsection{Adversarial Exit-wise Orthogonal Knowledge Distillation (AEOKD)}

\textbf{Overall NEO-KD loss.} Finally, considering the proposed loss functions in Eq. \eqref{equation: NKD}, \eqref{eq: EOKD} and the  original adversarial training loss, the overall objective function of our scheme  is written as follows:
\begin{multline}\label{eq: NEO-KD}
\mathcal{L}=\frac{1}{N}\sum_{j=1}^{N}\sum_{i=1}^{L}[\ell(f_{\theta_{i}}(x_j),y_j)+\ell(f_{\theta_{i}}(x^{adv}_j),y_j)+\gamma_i  (\alpha\cdot  NKD_{i,j}	+\beta\cdot EOKD_{i,j})],
\end{multline}
where $\alpha, \beta$ control the weights for each component and $\gamma_i$ is the knowledge distillation weight for each exit $i$. Since later exits have lower knowledge distillation loss than the earlier exits, we impose a slightly higher $\gamma_i$ for the later exits than $\gamma_i$ of the earlier exits. More details regarding hyperparameters are described in  Appendix. 

By introducing two unique components - NKD and EOKD - the overall NEO-KD loss function in Eq. \eqref{eq: NEO-KD}   reduces  adversarial transferability in the multi-exit network while correctly  guiding the output of the adversarial examples in each exit, significantly improving the adversarial robustness of multi-exit networks, as we will see in the next section.

%\textcolor{red}{***************OK until here***************}

%Since AKD and EOKD are distillation based scheme, the methods are fine-tuned by distillation parameter, $\alpha$ and $\beta$.
%Also, since later exits have lower knowledge distillation loss than earlier exits, we give slightly higher $\gamma_i$ for later exits than $\gamma_i$ for earlier exits.
%Equation \ref{eq: AEOKD} describes AEOKD loss function which is combined with loss function for adversarial training, AKD, and EOKD.
%The proposed AEOKD objective function sums up the losses of all exits and averages the losses of all data. 

\vspace{-1mm}

\section{Experiments}\label{sec:exp}
\vspace{-1mm}
%\subsubsection{Experiment Details}
In this section, we evaluate our method on five   datasets commonly adopted in multi-exit networks: MNIST \cite{lecun1998gradient}, CIFAR-10, CIFAR-100 \cite{krizhevsky2009learning},  Tiny-ImageNet \cite{le2015tiny}, and ImageNet \cite{russakovsky2015imagenet}.  For MNIST, we use SmallCNN \cite{hu2019triple} with 3 exits. We trained the MSDNet \cite{huang2017multi} with 3 and 7 exits using CIFAR-10 and CIFAR-100, respectively. For Tiny-ImageNet and ImageNet,  we trained the MSDNet with 5 exits. More  implementation details are provided in  Appendix.
\vspace{-1mm}

\subsection{Experimental Setup}
\vspace{-1mm}
%We train a $7$-exit MSDNet \cite{huang2017multi} for $150$ epochs with batch size $256$ of CIFAR100 dataset, $5$-exit MSDNet for $150$ epochs with batch size $256$ of CIFAR10 dataset, and $3$-exit SmallCNN \cite{hu2019triple} for $60$ epochs with batchsize $256$ of MNIST dataset.
%Other settings for MSDNet follow \cite{huang2017multi}\footnote{https://github.com/kalviny/MSDNet-PyTorch} except for channel which is $8$.
%Also, other settings for SmallCNN follow \cite{hu2019triple}\footnote{https://github.com/VITA-Group/triple-wins}.
%We use SGD optimizer with momentum 0.9 and weight decay $5\times10^{-4}$.
%Initial learning rate for CIFAR100 and CIFAR10 dataset is 0.1 and is decayed 10 fold at $75$-th epoch and $115$-th epoch.
%For MNIST dataset, we set initial learning rate 0.01 and the learning rate is decayed 10 fold at $30$-th epoch.

\textbf{Generating adversarial examples.}  To generate adversarial examples during training and testing, we utilize max-average attack and average attack proposed in \cite{hu2019triple}. 
%Consider a multi-exit network with $L$ exits. For the max-average attack, we first generate $L$ adversarial examples targeting each exit  via \textit{attacker algorithm} (e.g., FGSM, PGD), and then  select one adversarial example that  maximizes the sum  of losses from all exits. On the other hand, average attack generates an adversarial example that maximizes the average loss of all exits through the attacker algorithm. 
We perform adversarial training using adversarial examples generated by max-average attack, where the results for adversarial training via average attack are reported in Appendix. During training, we use PGD attack \cite{madry2018towards} with $7$ steps as attacker algorithm  to generate max-average and average attack while PGD attack with $50$ steps is adopted at test time for measuring robustness against a stronger attack. We further consider other strong attacks in Section \ref{subsec:ablation}.  
% Experiment results with average attack are in Appendix.
%Average attack seems more reasonable white box adversarial attack for multi-exit architecture as \cite{chen2022towards} is mentioned.  
%In other word, gradient to maximize averaged loss from all exits is used to FGSM method.
%For PGD attack, the step size $\alpha$ is $0.0078$ for CIFAR100 and CIFAR10 dataset, and $0.078$ for MNIST dataset. The number of iteration is commonly $10$. 
In each  attacker algorithm, the perturbation degree $\epsilon$ is $0.3$ for MNIST, and $8/255$ for CIFAR-10/100, and $2/255$ for Tiny-ImageNet/ImageNet datasets during adversarial training and when measuring the adversarial test accuracy.
%$w_i$ is $[1,1,1,1.5,1.5,1.5,1.5]$ for CIFAR100 dataset, $[1,1,1,1.5,1.5]$ for CIFAR10 dataset, and $[1,1,1]$ for MNIST dataset. 
%$\alpha, \beta$ value is $0.2, 0.5$ for all datasets.
Other details for generating adversarial examples and additional experiments on various  attacker algorithms are described in Appendix.

\textbf{Evaluation metric.} 
We evaluate the adversarial test accuracy as in \cite{hu2019triple}, which is the classification accuracy on the corrupted test dataset compromised by an attacker (e.g., average attack). We also measure the clean test accuracy  using the original clean test data and report the results in Appendix.

%\subsection{Experiment Results}
%We adversarially trained MSDNet \cite{huang2017multi} against max-average attack with CIFAR10, CIFAR100, and MNIST dataset (Table \ref{table: MaxAvg AdvT CIFAR10}, \ref{table: MaxAvg AdvT},  \ref{table: MaxAvg AdvT MNIST}).
\textbf{Baselines.} We compare our NEO-KD   with the following baselines. First, we consider the scheme based on adversarial training  without any knowledge distillation, where adversarial examples are generated by max-average attack or average attack \cite{hu2019triple}. The second baseline is the conventional self-knowledge distillation (SKD) strategy  \cite{li2019improved, phuong2019distillation} combined with adversarial training:  during adversarial training,   the prediction of the last exit for a given clean/adversarial data is distilled to the predictions of all the previous exits for the same clean/adversarial data.  
The third baseline is the knowledge distillation scheme for adversarial training \cite{goldblum2020adversarially}, which distills the prediction of clean data to the prediction of adversarial examples in  single-exit networks. As in \cite{goldblum2020adversarially}, we distill the last output of clean data to the last output of adversarial data.
The last baseline is a regularizer-based adversarial training strategy for multi-exit networks \cite{chen2022towards}, where the regularizer restricts the weights of the
fully connected layers (classifiers). In Appendix we compare our NEO-KD with the recent work TEAT \cite{dong2021exploring}, a general defense algorithm for single-exit networks.
%\textcolor{red}{For our scheme, we compare the adversarial test accuracy of our NEO-KD based on adversarial training with max-average attack with the baselines to confirm the effectiveness of the proposed method.}  

%\textcolor{blue}{Built upon the adversarial training via max-average attack, we apply our distillation-based robust training strategies, AKD, EOKD and the unified scheme combining them, denoted as AEOKD. We compare the adverarial test accuracy of our method with the baselines to confirm the effectiveness of the proposed methods.}

\begin{table*}[!t]
 	\tiny
	\centering
	\subfloat[Max-average attack\label{table: MaxAvg AdvT MaxAvg Attack MNIST}]{
		\begin{subtable}{0.48\linewidth}
			\centering
			\begin{tabular}{c|ccc||c}
				\hline
				Exit & 1 & 2 & 3 & Average\\
				\hline
				Adv.   w/o Distill  \cite{hu2019triple} & 89.74\% & 95.89\% & 96.82\% & 94.15\% \\
				SKD \cite{phuong2019distillation} & 89.77\% & 96.24\% & \textbf{97.08\%} & 94.36\% \\
				ARD \cite{goldblum2020adversarially} & 89.65\% & 95.79\% & 96.47\% & 93.97\% \\
				LW \cite{chen2022towards} & 87.08\% & 93.86\% & 95.42\% & 92.12\% \\
				\hline
				NEO-KD (ours) & \textbf{90.56\%} & \textbf{96.30\%} & 96.62\% & \textbf{94.49\%} \\
				\hline
			\end{tabular}
			%\caption{}
		\end{subtable}
	}
	%\hfill
	%\newline
	%\caption{Adversarial Test Accuracy against Max-Average Attack for Anytime Prediction Setting. All methods in this table are built upon adversarial training with max-average attack.}
	\subfloat[Average attack\label{table: MaxAvg AdvT Avg Attack MNIST}]{
		\begin{subtable}{0.48\linewidth}
			\centering
			\tiny
			\begin{tabular}{c|ccc||c}
				\hline
				Exit & 1 & 2 & 3 & Average\\
				\hline
				Adv.   w/o Distill  \cite{hu2019triple} & 85.17\% & 95.87\% & 97.20\% & 92.75\% \\
				SKD \cite{phuong2019distillation} & 84.35\% & 96.53\% & \textbf{97.53\%} & 92.82\% \\
				ARD \cite{goldblum2020adversarially} & 85.07\% & 96.12\% & 97.28\% & 92.82\% \\
				LW \cite{chen2022towards} & 85.26\% & 94.54\% & 96.05\% & 91.95\% \\
				\hline
				NEO-KD (ours) & \textbf{86.17\%} & \textbf{96.92\%} & 97.42\% & \textbf{93.50\%} \\
				\hline
			\end{tabular}
			%caption{Average attack}
			%label{table: MaxAvg AdvT Avg Attack MNIST}
		\end{subtable}
	}
	%	\hfill
	%	\newline
	%	\begin{subtable}[!t]{1\linewidth}
		%		\centering
		%		\begin{tabular}{c|cccccc}
			%			Exit & Standard & MaxAvg & SKD & AKD & EOKD & AEOKD\\
			%			& \cite{huang2017multi} & \cite{hu2019triple} & \cite{phuong2019distillation} & (ours) & (ours) & (ours)\\
			%			\hline
			%			1 & 99.448\% & 98.367\% & 99.625\% & \textbf{98.452\%} & 98.337\% & 98.403\%\\
			%			2 & 99.922\% & 99.785\% & 99.982\% & 99.798\% & 99.795\% & \textbf{99.810\%}\\
			%			3 & 99.972\% & 99.967\% & 99.993\% & 99.958\% & 99.970\% & \textbf{99.970\%}\\
			%			\hline
			%			Average & 99.781\% & 99.373\% & 99.867\% & \textbf{99.403\%} & 99.367\% & 99.394\%\\
			%		\end{tabular}
		%		\caption{Clean Test Accuracy}
		%		\label{table: MaxAvg AdvT clean test accuracy MNIST}
		%	\end{subtable}
	\vspace{-1mm}
	\caption{\small \textbf{Anytime prediction setup}: Adversarial test accuracy on MNIST. }
	\label{table: MaxAvg AdvT MNIST}
	\vspace{-2mm}
\end{table*}

\begin{table*}[!t]
	\centering
	\subfloat[Max-average attack\label{table: MaxAvg AdvT MaxAvg Attack CIFAR10}]{
		\begin{subtable}[!t]{0.48\linewidth}
			\centering
			\tiny
			\begin{tabular}{c|ccc||c}
				\hline
				Exit & 1 & 2 & 3 & Average\\
				\hline
				Adv.   w/o Distill  \cite{hu2019triple} & 44.77\% & 46.10\% & 46.86\% & 45.91\% \\
				SKD \cite{phuong2019distillation} & 44.75\% & 44.54\% & 44.79\% & 44.69\% \\
				ARD \cite{goldblum2020adversarially} & 44.50\% & 45.85\% & \textbf{51.82\%} & 47.39\% \\
				LW \cite{chen2022towards} & 37.38\% & 35.39\% & 34.85\% & 35.87\% \\
				\hline
				NEO-KD (ours) & \textbf{46.53\%} & \textbf{47.65\%} & 50.71\% & \textbf{48.30\%} \\
				\hline
			\end{tabular}
		\end{subtable}
	}
	%\caption{Adversarial Test Accuracy against Max-Average Attack for Anytime Prediction Setting. All methods in this table are built upon adversarial training with max-average attack.}
	\subfloat[Average attack\label{table: MaxAvg AdvT Avg Attack CIFAR10}]{
		\begin{subtable}[!t]{0.48\linewidth}
			\centering
			\tiny
			\begin{tabular}{c|ccc||c}
				\hline
				Exit & 1 & 2 & 3 & Average\\
				\hline
				Adv.   w/o Distill  \cite{hu2019triple} & 38.00\% & 41.42\% & 40.70\% & 40.04\% \\
				SKD \cite{phuong2019distillation} & 39.36\% & 41.39\% & 38.39\% & 39.71\% \\
				ARD \cite{goldblum2020adversarially} & 39.37\% & 41.98\% & 43.53\% & 41.63\% \\
				LW \cite{chen2022towards} & 31.47\% & 31.41\% & 28.98\% & 30.62\% \\
				\hline
				NEO-KD (ours) & \textbf{41.67\%} & \textbf{45.38\%} & \textbf{45.54\%} & \textbf{44.20\%} \\
				\hline
			\end{tabular}
		\end{subtable}
	}
	%	\hfill
	%	\newline
	%	\begin{subtable}[!t]{1\linewidth}
		%		\centering
		%		\begin{tabular}{c|cccccc}
			%			Exit & Standard & MaxAvg & SKD & AKD & EOKD & AEOKD\\
			%			& \cite{huang2017multi} & \cite{hu2019triple} & \cite{phuong2019distillation} & (ours) & (ours) & (ours)\\
			%			\hline
			%			1 & 99.448\% & 98.367\% & 99.625\% & \textbf{98.452\%} & 98.337\% & 98.403\%\\
			%			2 & 99.922\% & 99.785\% & 99.982\% & 99.798\% & 99.795\% & \textbf{99.810\%}\\
			%			3 & 99.972\% & 99.967\% & 99.993\% & 99.958\% & 99.970\% & \textbf{99.970\%}\\
			%			\hline
			%			Average & 99.781\% & 99.373\% & 99.867\% & \textbf{99.403\%} & 99.367\% & 99.394\%\\
			%		\end{tabular}
		%		\caption{Clean Test Accuracy}
		%		\label{table: MaxAvg AdvT clean test accuracy MNIST}
		%	\end{subtable}
	\vspace{-1mm}
	\caption{\small \textbf{Anytime prediction setup}: Adversarial test accuracy on CIFAR-10. }
	\label{table: MaxAvg AdvT CIFAR10}
		\vspace{-2mm}
\end{table*}
\textbf{Inference scenarios.} At inference time, we consider two widely known setups for multi-exit networks: (i) anytime prediction setup and (ii) budgeted prediction setup. In the anytime prediction setup, an appropriate exit is selected   depending on the current latency constraint. In this setup, for each exit, we report the average performance computed with  all test samples.
In the budgeted prediction setup, given a fixed computational budget, each sample is predicted at different exits depending on the predetermined confidence threshold (which is determined by validation set). Starting from the first exit,  given a test sample, when the confidence at the exit (defined as the maximum softmax value) is larger than the threshold, prediction is made at this exit. Otherwise, the sample proceeds to the next exit.  In this scenario,  easier samples are predicted at earlier exits and harder samples are predicted at later exits, which leads to efficient inference. 
Given the fixed computation budget and confidence threshold, we measure the average accuracy of the test samples. %We evaluate our method in these two settings and will show that our method outperforms other baselines in both settings. 
 We evaluate our method in these two
settings and show that our method outperforms the baselines in both settings. 
 More detailed settings for our inference scenarios are provided in Appendix.
%\vspace{-1mm}

\begin{table*}[!t]

	\scriptsize
	\setlength{\tabcolsep}{2.5pt}
	\centering
	\begin{subtable}[!t]{1\linewidth}
		\centering
		\begin{tabular}{c|ccccccc|c|| ccccccc|c}
			\hline
			& \multicolumn{8}{c||}{top-1 accuracy (\%)} & \multicolumn{8}{c}{top-5 accuracy (\%)} \\ \hline
			Exit & 1 & 2 & 3 & 4 & 5 & 6 & 7 & Avg. & 1 & 2 & 3 & 4 & 5 & 6 & 7 & Avg.\\
			\hline
			Adv.   w/o Distill  \cite{hu2019triple} & 28.04  & 28.32 & 28.34 & 27.64 & 26.78  & 25.93 & 24.77 & 27.12 & \textbf{60.64} & \textbf{61.15} & 59.13 & 58.46 & 58.38 & 57.46 & 57.44 & 58.95 \\
			SKD \cite{phuong2019distillation} & 27.36 & 27.68 & 25.79 & 24.74 & 24.10 & 20.86 & 19.28 & 24.26 & 60.50 & 60.23 & 57.56 & 55.98 & 55.11 & 52.71 & 52.17 & 56.32 \\
			ARD \cite{goldblum2020adversarially} & 27.60  & 28.00 & 27.99  & 27.42 & 28.32  & 27.25 & 29.08  & 27.95 & 60.26  & 60.51 & 59.31 & 58.40  & 58.40 & 58.09 & \textbf{60.62} & 59.37 \\
			LW \cite{chen2022towards} & 20.44  & 20.57 & 19.86  & 19.34 & 19.51 & 19.46 & 20.22  & 19.91 & 50.69  & 50.91 & 48.47 & 48.56 & 48.13 & 49.61 & 49.76 & 49.45 \\
			\hline 
			NEO-KD (ours) & \textbf{28.37}  & \textbf{28.78} & \textbf{29.02} & \textbf{29.49} & \textbf{30.06} & \textbf{28.45} & \textbf{28.54} & \textbf{28.96} & 59.58 & 60.04 & \textbf{59.67} & \textbf{59.29} & \textbf{60.46} & \textbf{58.96} & 59.44 & \textbf{59.63} \\
			\hline
		\end{tabular}
		\caption{Max-average atack}
		\label{table: MaxAvg AdvT MaxAvg Attack}
	\end{subtable}
	\hfill
	\newline
	%\caption{Adversarial Test Accuracy against Max-Average Attack for Anytime Prediction Setting. All methods in this table are built upon adversarial training with max-average attack.}
	\begin{subtable}[!t]{1\linewidth}
		\scriptsize
		\setlength{\tabcolsep}{2.5pt}
		\centering
		\begin{tabular}{c|ccccccc|c|| ccccccc|c}
			\hline
			& \multicolumn{8}{c||}{top-1 accuracy (\%)} & \multicolumn{8}{c}{top-5 accuracy (\%)} \\ \hline
			Exit & 1 & 2 & 3 & 4 & 5 & 6 & 7 & Avg. & 1 & 2 & 3 & 4 & 5 & 6 & 7 & Avg.\\
			\hline
			Adv.   w/o Distill  \cite{hu2019triple} & 16.74  & 17.33 & 19.05 & 19.47 & 19.06 & 18.15 & 17.12 & 18.13 & 47.31 & 48.34 & 50.78 & 50.75 & 50.80 & 49.89 & 47.10 & 49.28 \\
			SKD \cite{phuong2019distillation} & 18.13 & 18.45 & 19.53 & 19.87 & 19.54 & 16.67 & 14.21 & 18.06 & \textbf{49.85} & 50.72 & 51.17 & 51.91 & 51.73 & 47.32 & 43.18 & 49.41 \\
			ARD \cite{goldblum2020adversarially} & 16.63  & 17.16 & 19.13  & 19.74 & 19.70 & 18.92 & 19.83  & 18.73 & 46.90 & 48.60 & 50.44 & 51.28 & 50.93 & 49.14 & 49.01 & 49.47 \\
			LW \cite{chen2022towards} & 13.78 & 13.95 & 15.16 & 15.62 & 15.86 & 13.01 & 13.57 & 14.42 & 41.14  & 41.27 & 43.29 & 43.67 & 44.08 & 40.49 & 40.66 & 42.09 \\
			\hline 
			NEO-KD (ours) & \textbf{20.41}  & \textbf{21.32} & \textbf{23.27} & \textbf{24.30} & \textbf{24.47} & \textbf{23.99} & \textbf{22.39} & \textbf{22.88} & 49.48 & \textbf{51.47} & \textbf{53.13} & \textbf{55.01} & \textbf{54.69} & \textbf{54.32} & \textbf{52.11} & \textbf{52.89} \\
			\hline
		\end{tabular}
		\caption{Average attack}
		\label{table: MaxAvg AdvT Avg Attack}
	\end{subtable}
	%	\hfill
	%	\newline
	%	\begin{subtable}[!t]{1\linewidth}
		%		\centering
		%		\begin{tabular}{c|cccccc}
			%			Exit & Standard & MaxAvg & SKD & AKD & EOKD & AEOKD\\
			%			& \cite{huang2017multi} & \cite{hu2019triple} & \cite{phuong2019distillation} & (ours) & (ours) & (ours)\\
			%			\hline
			%			1 & 63.23\% & 58.46\% & 16.36\% & \textbf{59.27\%} & 58.21\% & 58.79\%\\
			%			2 & 67.94\% & 62.62\% & 32.60\% & \textbf{62.87\%} & 62.17\% & 61.41\%\\
			%			3 & 69.73\% & 64.93\% & 35.74\% & \textbf{65.95\%} & 64.19\% & 64.25\%\\
			%			4 & 71.71\% & 67.25\% & 38.01\% & \textbf{67.40\%} & 66.49\% & 67.04\%\\
			%			5 & 72.71\% & 68.30\% & 34.66\% & \textbf{69.41\%} & 67.37\% & 68.41\%\\
			%			6 & 73.06\% & 69.28\% & 38.06\% & \textbf{70.22\%} & 68.52\% & 68.92\%\\
			%			7 & 73.50\% & 68.98\% & 38.87\% & \textbf{70.01\%} & 68.96\% & 69.53\%\\
			%			\hline
			%			Average & 70.27\% & 65.69\% & 33.47\% & \textbf{66.45\%} & 65.13\% & 65.48\%\\
			%		\end{tabular}
		%		\caption{Clean Test Accuracy}
		%		\label{table: MaxAvg AdvT clean test accuracy}
		%	\end{subtable}
\vspace{-1mm}
	\caption{\small \textbf{Anytime prediction setup}: Adversarial test accuracy on CIFAR-100. }
	\label{table: MaxAvg AdvT}
	\vspace{-2mm}
\end{table*}

\begin{table*}[!t]
	\scriptsize	
	\centering
	\begin{subtable}[!t]{1\linewidth}
		\setlength{\tabcolsep}{4.5pt}
		\centering
		\begin{tabular}{c|ccccc|c||ccccc|c}
			\hline
			& \multicolumn{6}{c||}{top-1 accuracy (\%)} & \multicolumn{6}{c}{top-5 accuracy (\%)} \\ \hline
			Exit & 1 & 2 & 3 & 4 & 5 & Avg. & 1 & 2 & 3 & 4 & 5 & Avg.\\
			\hline
			Adv.   w/o Distill  \cite{hu2019triple} & 31.30 & 32.38 & 32.52 & 31.42 & 31.56 & 31.84 & 60.20 & 60.84 & 61.10 & 58.64 & 59.56 & 60.07 \\
			SKD \cite{phuong2019distillation} & \textbf{33.04} & 33.24 & 30.50 & 28.40 & 28.50 & 30.74 & \textbf{63.68} & 62.72 & 60.04 & 58.50 & 58.28 & 60.64 \\
			ARD \cite{goldblum2020adversarially} & 30.08 & 31.84 & 30.56  & 31.22 & 32.34 & 31.21 & 59.88 & 60.74 & 59.62 & 59.28 & 59.72 & 59.85 \\
			LW \cite{chen2022towards} & 26.06 & 26.00 & 24.54 & 24.42 & 24.88 & 25.18 & 53.24 & 51.90 & 51.02  & 51.02 & 51.14 & 51.66 \\
			\hline 
			NEO-KD (ours) & 32.96  & \textbf{35.08} & \textbf{33.42} & \textbf{32.40} & \textbf{32.64} & \textbf{33.30} & 62.48 & \textbf{62.82} & \textbf{61.76} & \textbf{60.70} & \textbf{60.92} & \textbf{61.74} \\
			\hline
		\end{tabular}
		\caption{ Max-average attack}
		\label{table: MaxAvg AdvT MaxAvg Attack Tiny-ImageNet}
	\end{subtable}
	\hfill
	\newline
	%\caption{Adversarial Test Accuracy against Max-Average Attack for Anytime Prediction Setting. All methods in this table are built upon adversarial training with max-average attack.}
	\begin{subtable}[!t]{1\linewidth}
		\setlength{\tabcolsep}{4.5pt}
		\centering
		\begin{tabular}{c|ccccc|c||ccccc|c}
			\hline
			& \multicolumn{6}{c||}{top-1 accuracy (\%)} & \multicolumn{6}{c}{top-5 accuracy (\%)} \\ \hline
			Exit & 1 & 2 & 3 & 4 & 5 & Avg. & 1 & 2 & 3 & 4 & 5 & Avg.\\
			\hline
			Adv.   w/o Distill  \cite{hu2019triple} & 25.22 & 27.34 & 28.94 & 28.06 & 28.48 & 27.61 & 53.68 & 55.46 & 58.18 & 57.46 & 57.38 & 56.43 \\
			SKD \cite{phuong2019distillation} & \textbf{28.40} & 29.26 & 28.74 & 28.14 & 26.82 & 28.27 & \textbf{58.34} & 59.30 & 58.38 & 57.72 & 55.48 & 57.84 \\
			ARD \cite{goldblum2020adversarially} & 24.48 & 26.76 & 27.78  & 28.46 & 29.14 & 27.32 & 53.44 & 56.10 & 57.20 & 57.32 & 57.08 & 56.23 \\
			LW \cite{chen2022towards} & 22.34 & 23.12 & 23.58 & 22.76 & 23.30 & 23.02 & 47.56 & 48.16 & 49.86 & 48.64 & 49.72 & 48.79 \\
			\hline 
			NEO-KD (ours) & 28.24  & \textbf{31.14} & \textbf{30.58} & \textbf{31.58} & \textbf{31.24} & \textbf{30.56} & 57.34 & \textbf{59.82} & \textbf{60.16} & \textbf{60.04} & \textbf{59.06} & \textbf{59.28} \\
			\hline
		\end{tabular}
		\caption{Average attack}
		\label{table: MaxAvg AdvT Avg Attack Tiny-ImageNet}
	\end{subtable}
	%	\hfill
	%	\newline
	%	\begin{subtable}[!t]{1\linewidth}
		%		\centering
		%		\begin{tabular}{c|cccccc}
			%			Exit & Standard & MaxAvg & SKD & AKD & EOKD & AEOKD\\
			%			& \cite{huang2017multi} & \cite{hu2019triple} & \cite{phuong2019distillation} & (ours) & (ours) & (ours)\\
			%			\hline
			%			1 & 90.01\% & 86.78\% & 86.86\% & 86.49\% & 85.55\% & \textbf{86.90\%}\\
			%			2 & 91.18\% & \textbf{88.95\%} & 89.07\% & 88.93\% & 87.45\% & 88.38\%\\
			%			3 & 91.97\% & 90.16\% & 90.16\% & \textbf{90.24\%} & 89.25\% & 89.42\%\\
			%			4 & 92.54\% & \textbf{91.05\%} & 91.00\% & 90.75\% & 90.43\% & 90.34\%\\
			%			5 & 92.35\% & \textbf{91.23\%} & 91.38\% & 91.05\% & 90.96\% & 90.79\%\\
			%			\hline
			%			Average & 91.61\% & \textbf{89.63\%} & 89.69\% & 89.49\% & 88.73\% & 89.17\%\\
			%		\end{tabular}
		%		\caption{Clean Test Accuracy}
		%		\label{table: MaxAvg AdvT clean test accuracy CIFAR10}
		%	\end{subtable}
\vspace{-1mm}
	\caption{\small \textbf{Anytime prediction setup}: Adversarial test accuracy on Tiny-ImageNet. }
	\label{table: MaxAvg AdvT Tiny-ImageNet}
		\vspace{-2mm}
\end{table*}

\begin{table*}[!t]
	\scriptsize	
	\centering
	%\caption{Adversarial Test Accuracy against Max-Average Attack for Anytime Prediction Setting. All methods in this table are built upon adversarial training with max-average attack.}
	\begin{subtable}[!t]{1\linewidth}
		\setlength{\tabcolsep}{4.5pt}
		\centering
		\begin{tabular}{c|ccccc|c||ccccc|c}
			\hline
			& \multicolumn{6}{c||}{top-1 accuracy (\%)} & \multicolumn{6}{c}{top-5 accuracy (\%)} \\ \hline
			Exit & 1 & 2 & 3 & 4 & 5 & Avg. & 1 & 2 & 3 & 4 & 5 & Avg.\\
			\hline
			Adv.   w/o Distill  \cite{hu2019triple} & 26.10 & 31.71 & 31.94 & 30.22 & 32.30 & 30.45 & 55.40 & 59.84 & 59.23 & 57.17 & 59.90 & 58.31 \\
			SKD \cite{phuong2019distillation} & 27.34 & 31.44 & 29.63 & 27.03 & 26.71 & 28.43 & \textbf{56.96} & \textbf{59.94} & 57.49 & 53.98 & 55.24 & 56.72 \\
			ARD \cite{goldblum2020adversarially} & 25.81 & 32.00 & 32.93 & 32.00 & 31.72 & 30.89 & 54.46 & 59.34 & 59.24 & 57.93 & 58.29 & 57.85\\
			\hline 
			NEO-KD (ours) & \textbf{27.89} & \textbf{32.61} & \textbf{32.99} & \textbf{32.74} & \textbf{35.63} & \textbf{32.37} & 55.46 & 59.64 & \textbf{59.79} & \textbf{59.61} & \textbf{62.46} & \textbf{59.39} \\
			\hline
		\end{tabular}
		\caption{Max-average attack}
	\end{subtable}
	\begin{subtable}[!t]{1\linewidth}
	\setlength{\tabcolsep}{4.5pt}
	\centering
	\begin{tabular}{c|ccccc|c||ccccc|c}
		\hline
		& \multicolumn{6}{c||}{top-1 accuracy (\%)} & \multicolumn{6}{c}{top-5 accuracy (\%)} \\ \hline
		Exit & 1 & 2 & 3 & 4 & 5 & Avg. & 1 & 2 & 3 & 4 & 5 & Avg.\\
		\hline
		Adv.   w/o Distill  \cite{hu2019triple} & 18.20 & 24.21 & 28.24 & 28.73 & 28.50 & 25.58 & 44.74 & 52.57 & 56.82 & 57.55 & 57.14 & 53.77 \\
		SKD \cite{phuong2019distillation} & 19.04 & 24.45 & 26.45 & 25.72 & 22.17 & 23.57 & 46.31 & 52.92 & 55.33 & 54.71 & 50.18 & 51.89 \\
		ARD \cite{goldblum2020adversarially} & 17.56 & 23.68 & 27.98 & 28.52 & 25.83 & 24.71 & 43.76 & 52.07 & 56.42 & 57.30 & 53.48 & 52.60\\
		\hline 
		NEO-KD (ours) & \textbf{22.42} & \textbf{28.62} & \textbf{31.77} & \textbf{32.78} & \textbf{34.30} & \textbf{29.98} & \textbf{48.02} & \textbf{55.30} & \textbf{58.76} & \textbf{60.07} & \textbf{61.35} & \textbf{56.70} \\
		\hline
	\end{tabular}
	\caption{Average attack}
	\end{subtable}
	\vspace{-1mm}
	\caption{\small \textbf{Anytime prediction setup}: Adversarial test accuracy on ImageNet. }
	\label{table: MaxAvg AdvT ImageNet}
	\vspace{-2mm}
\end{table*}

 	\vspace{-1mm}

\subsection{Main Experimental Results} \label{sec:main_exp}
 	\vspace{-1mm}

\textbf{Result 1: Anytime prediction setup.}
Tables  \ref{table: MaxAvg AdvT MNIST}, \ref{table: MaxAvg AdvT CIFAR10}, \ref{table: MaxAvg AdvT},  \ref{table: MaxAvg AdvT Tiny-ImageNet}, \ref{table: MaxAvg AdvT ImageNet} compare the  adversarial test accuracy of different schemes under max-average attack and average attack using MNIST, CIFAR-10/100, Tiny-ImageNet, and ImageNet, respectively. %The first observation drawn from the results is that the scheme without adversarial training is highly vulnerable to the adversarial examples generated by both attacks. Next, the performance of SKD generally achieves  lower performance  than  \textit{Adv. w/o Distill}. This means that naively applying SKD (which is the existing self-distillation technique) during adversarial training results in performance degradation, indicating the importance of knowledge distillation strategies for adversarial training in multi-exit networks.
% SKD is low performance, ARD is high performance --> the distillation method is important.
Note that we achieve adversarial accuracies between 40\% - 50\% for CIFAR-10, which is standard considering the prior works on robust multi-exit networks \cite{hu2019triple}. Our first observation from the results indicates that the performance of SKD \cite{phuong2019distillation} is generally lower than that of \textit{Adv. w/o Distill}, whereas ARD \cite{goldblum2020adversarially} outperforms \textit{Adv. w/o Distill}. This suggests
that the naive application of self-knowledge distillation can either  increase or decrease the adversarial robustness of multi-exit networks. Consequently, the method of knowledge distillation significantly influences the robustness of multi-exit networks (i.e., determining which knowledge to distill and which exit to target). To further enhance robustness, we investigate strategies for distilling high-quality knowledge and mitigating adversarial transferability.
By combining EOKD with NKD to mitigate dependency across submodels while guiding a multi-exit network to extract high quality features from adversarial examples as original data, NEO-KD achieves the highest adversarial test accuracy at most exits  compared to the baselines   for all  datasets/attacks.  The overall results confirm the  advantage of  NEO-KD for robust multi-exit networks.
%These experimental results prove that  distilling the ensembled output of neighbor exits of clean data to adversarial examples, and reducing dependency across exits are effective to improve robustness of multi-exit neural network. 
%; there is an implicit synergy when combining AKD and EOKD by balancing distillation losses.
%Note that in CIFAR-10, EOKD has lower adversarial test accuracy than the adversarial training without distillation,   while AKD shows comparable performance with this baseline. For CIFAR-10, each exit is provided only one non-maximal class  ($\lfloor (10-1)/5 \rfloor = 1$) according to the orthogonal label operation $O(\cdot)$ defined in \eqref{eq: EOKD}. On the other hand, EOKD provides $3$ non-maximal class in MNIST  ($\lfloor (10-1)/3 \rfloor = 3$) and $14$ non-maximal class in CIFAR-100  ($\lfloor (100-1)/7 \rfloor = 14$). Therefore, the distillation of knowledge from clean data to adversarial data in CIFAR-10 is somewhat limited when  EOKD is used solely, again indicating the importance of using AKD and EOKD together. 

%\vspace{-1mm}

\begin{figure*}[!t]
		\vspace{-3mm}
	\centering
	\begin{subfigure}{.33\textwidth}
		\centering
		\includegraphics[width=1\linewidth]{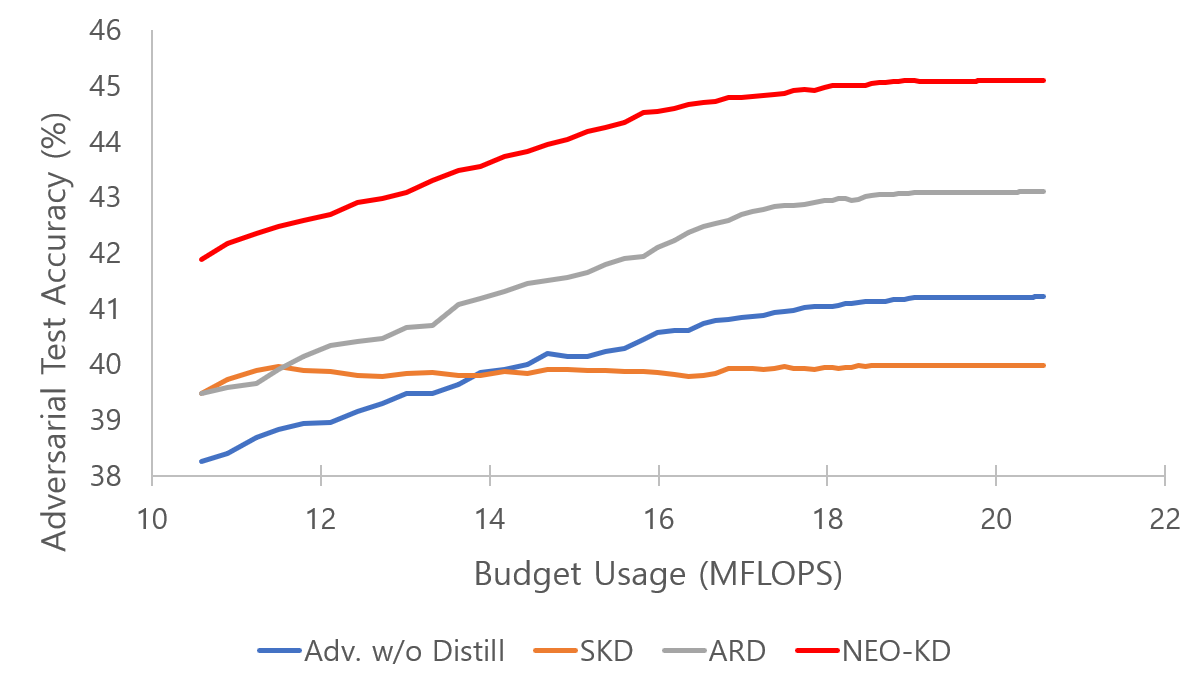}
		\caption{CIFAR-10}
		\label{figure: Adv Test Accuracy for Each Budget(MaxAvg, Avg Attack, CIFAR10)}
	\end{subfigure}
	\begin{subfigure}{.33\textwidth}
		\centering
		\includegraphics[width=1.0\linewidth]{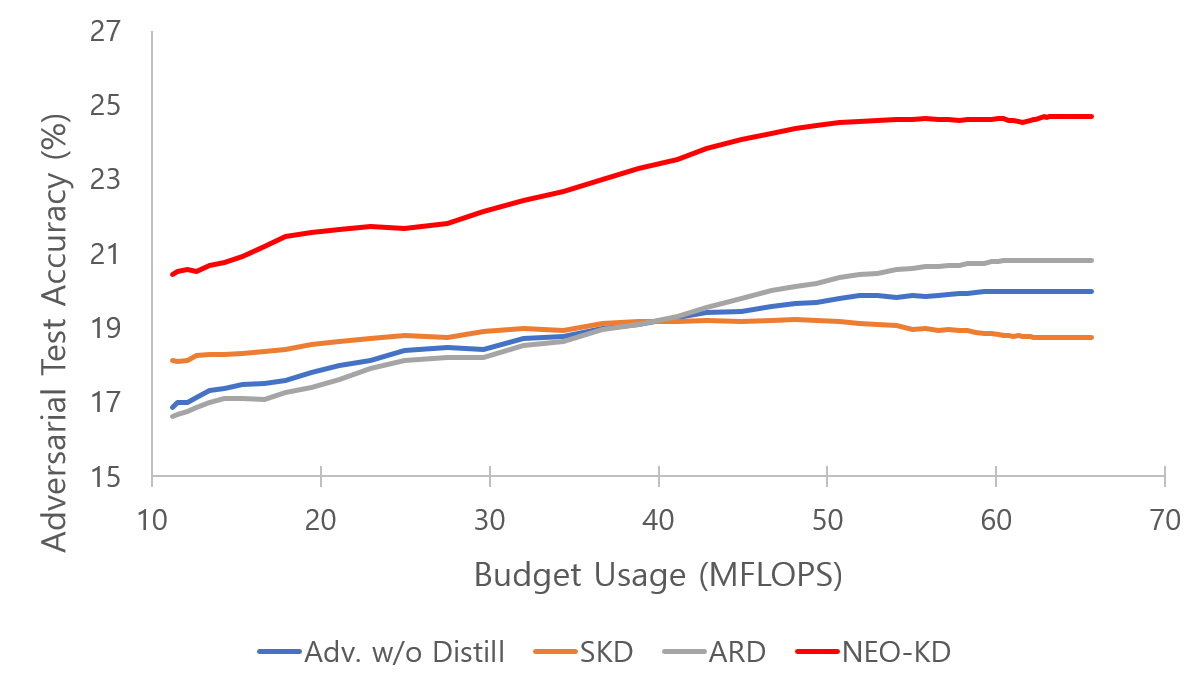}
		\caption{CIFAR-100}
		\label{figure: Adv Test Accuracy for Each Budget(MaxAvg, Avg Attack)}
	\end{subfigure}
	\begin{subfigure}{.3\textwidth}
		\centering
		\includegraphics[width=1\linewidth]{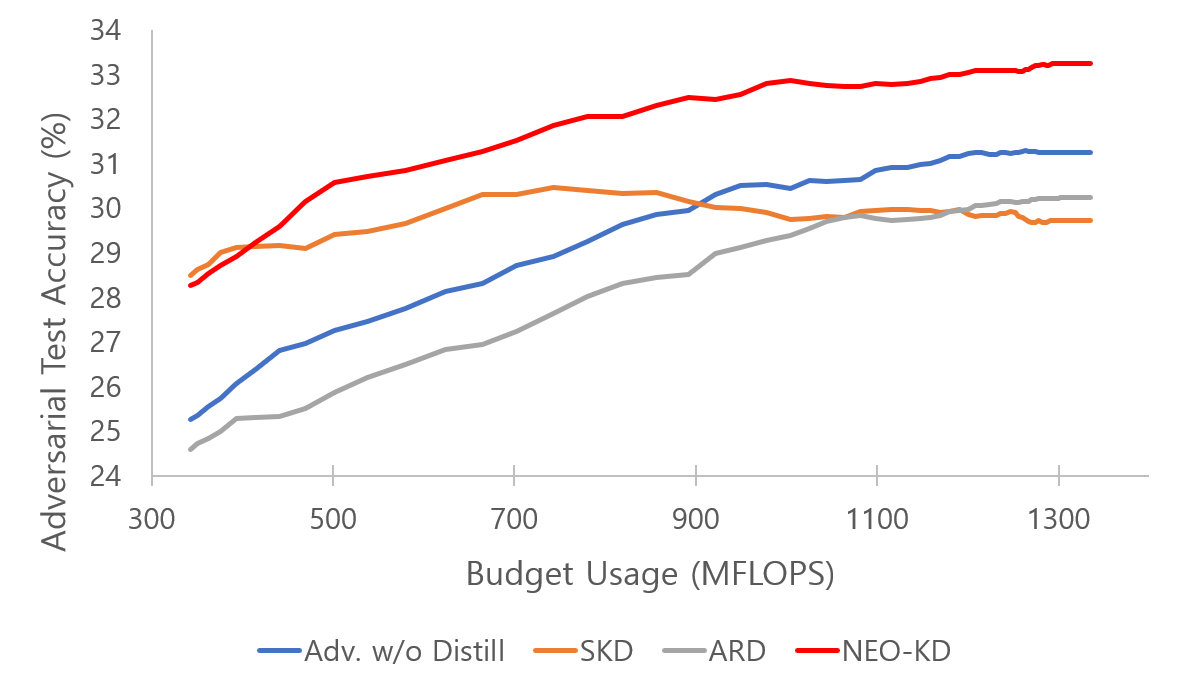}
		\caption{Tiny-ImageNet}
		\label{figure: Adv Test Accuracy for Each Budget(MaxAvg, Avg Attack, Tiny-ImageNet)}
	\end{subfigure}
	
	%	\begin{subfigure}{.33\textwidth}
		%		\centering
		%		\includegraphics[width=1\linewidth]{Clean Test Accuracy for Each Budget(MaxAvg).png}
		%		\caption{Clean Test Accuracy.}
		%		\label{figure: Clean Test Accuracy for Each Budget(MaxAvg)}
		%	\end{subfigure}
	\vspace{-1mm}
	\caption{\small \textbf{Budgeted prediction setup}: Adversarial test accuracy under average attack. The result for LW is excluded since the performance  is too low and thus hinders the  comparison between baselines and   NEO-KD.}
	\label{figure: Budget, Average Attack}
	\vspace{-4mm}
\end{figure*}

\begin{figure*}[!t]

	\centering
	\begin{subfigure}{.32\textwidth}
		\centering
		\includegraphics[width=1\linewidth]{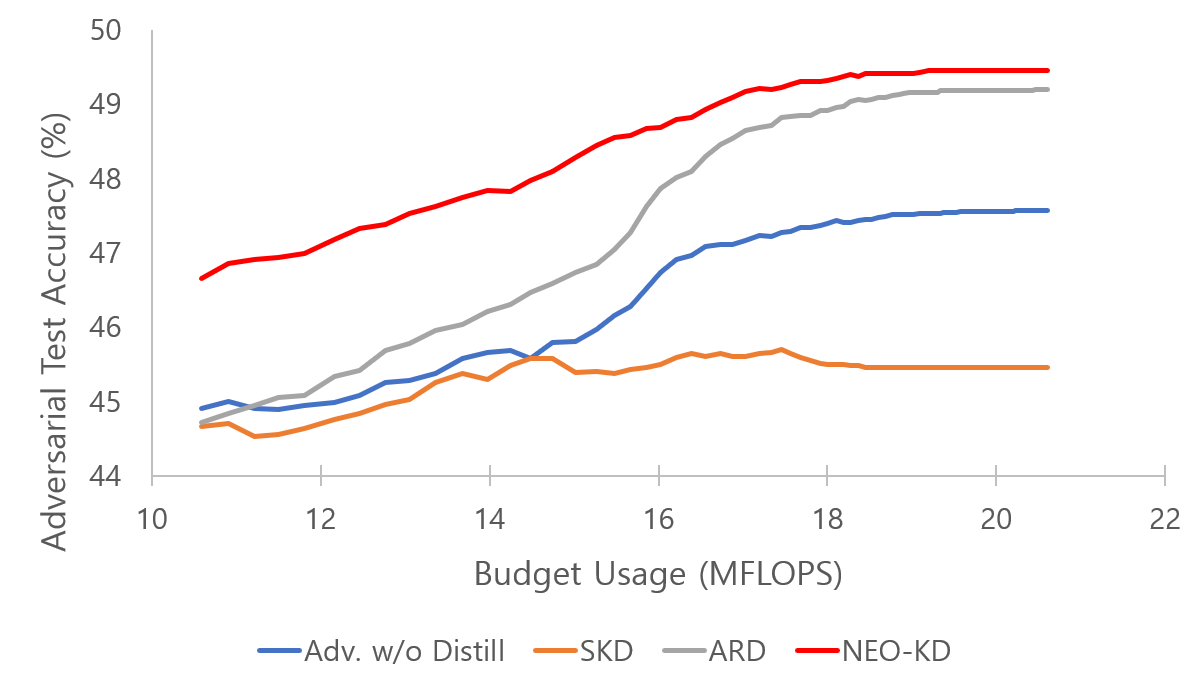}
		\caption{CIFAR-10}
		\label{figure: Adv Test Accuracy for Each Budget(MaxAvg, MaxAvg Attack, CIFAR10)}
	\end{subfigure}
	\begin{subfigure}{.32\textwidth}
		\centering
		\includegraphics[width=1.0\linewidth]{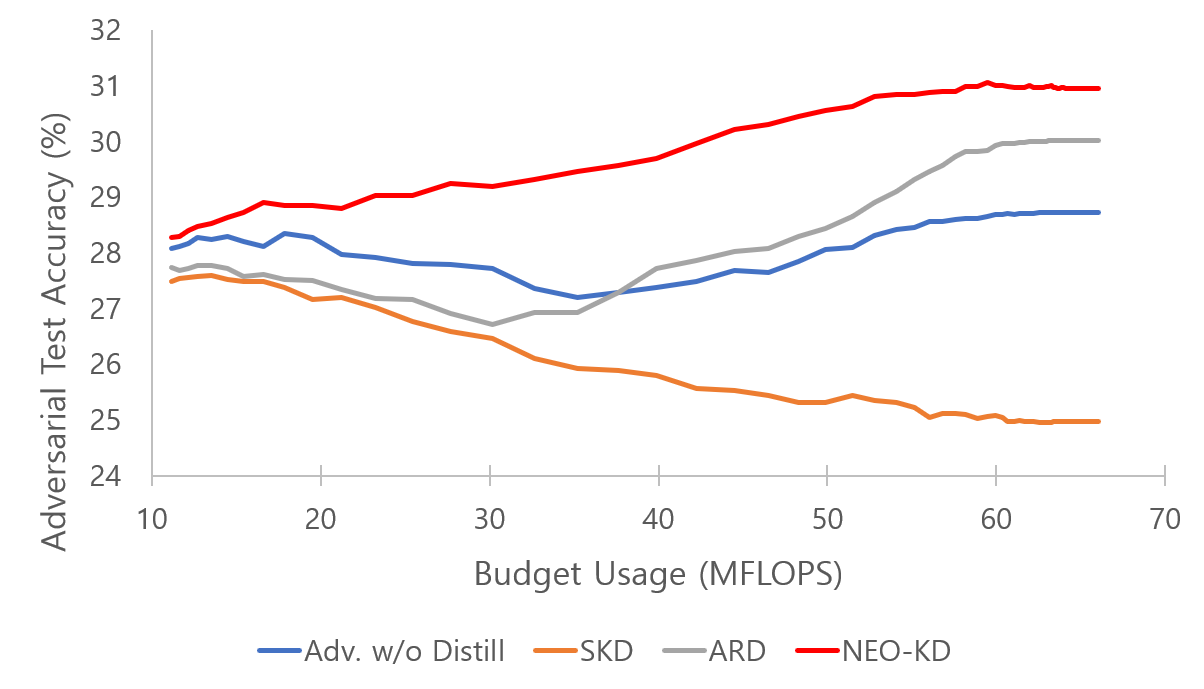}
		\caption{CIFAR-100}
		\label{figure: Adv Test Accuracy for Each Budget(MaxAvg, MaxAvg Attack)}
	\end{subfigure}
	\begin{subfigure}{.32\textwidth}
		\centering
		\includegraphics[width=1\linewidth]{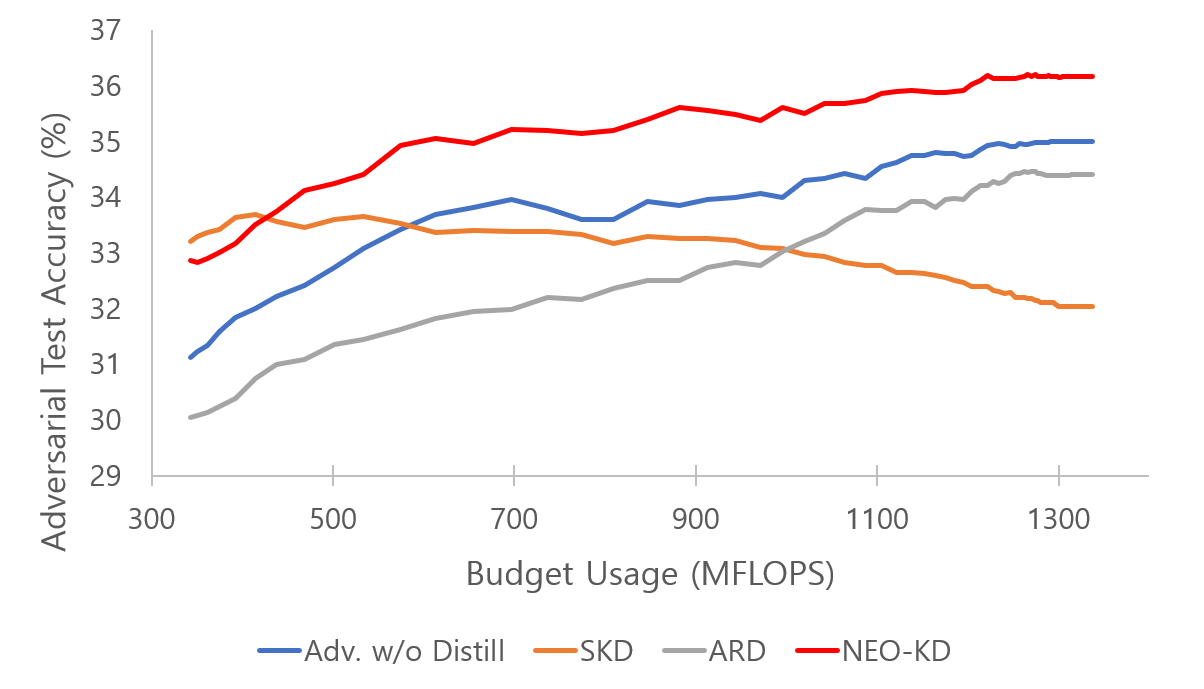}
		\caption{Tiny-ImageNet}
		\label{figure: Adv Test Accuracy for Each Budget(MaxAvg, MaxAvg Attack, Tiny-ImageNet)}
	\end{subfigure}
	
	%	\begin{subfigure}{.33\textwidth}
		%		\centering
		%		\includegraphics[width=1\linewidth]{Clean Test Accuracy for Each Budget(MaxAvg, CIFAR10).png}
		%		\caption{Clean Test Accuracy.}
		%		\label{figure: Clean Test Accuracy for Each Budget(MaxAvg, CIFAR10)}
		%	\end{subfigure}
	\vspace{-1mm}
	\caption{\small \textbf{Budgeted prediction setup}: Adversarial test accuracy under max-average attack. The result for LW is excluded since the performance  is too low and thus hinders the  comparison between baselines and  NEO-KD.}
	\label{figure: Budget, Max-Average Attack}
	\vspace{-4mm}
\end{figure*}

\textbf{Result 2: Budgeted prediction setup.}
%In budgeted prediction setup, a specific exit is selected depending on a given budget constraint. Specifically, based on the predetermined confidence threshold (through validation set), the modal save the computing budget for easy samples and utilize the saved budget to predict hard samples. The detailed descriptions about how to determine the confidence threshold is in Supplementary Material. 
Different from the anytime prediction setup where the pure performance of each exit is measured, in this setup, we adopt \textit{ensemble strategy} at inference time where the predictions from the selected exit (according to the confidence threshold) and the previous exits are ensembled. From the results in   anytime prediction setup, it is observed  that various  schemes tend to show low performance at the later exits compared to earlier exits in the model, where more details are discussed in  Appendix. Therefore, this \textit{ensemble strategy} can boost the performance of the later exits. With the ensemble scheme, given a fixed computation budget, we compare  adversarial test accuracies of our method with the baselines.

Figures \ref{figure: Budget, Average Attack} and \ref{figure: Budget, Max-Average Attack}  show the results in budgeted prediction  setup under average attack and max-average attack, respectively. %show the comparison of adversarial test accuracy against max-average attack and average attack when we train a model adversarially with max-average attack.
%By adopting ensemble scheme, the weak last exit can be compensated and the performance of later exits increase as we use lots of budget.
%If we do not adopt ensemble scheme, using more budget can be detrimental to the performance since the last exit has low performance.
%With the ensemble inference, the comparison of adversarial test accuracy shows similar aspect with anytime prediction setup.
NEO-KD achieves the best adversarial test accuracy against both average and max-average attacks in all budget setups. Our scheme also achieves the target accuracy with significantly smaller computing budget compared to the baselines.  For example, to achieve $41.21\%$ of accuracy against average attack using CIFAR-10 (which is the maximum accuracy of \textit{Adv. w/o Distill}),  the proposed NEO-KD needs  $10.59$ MFlops compared to Adv. w/o Distill that requires $20.46$ MFlops, saving  $48.24\%$ of computing budget. Compared to ARD, our NEO-KD saves $25.27\%$ of computation, while SKD and LW are unable to achieve this target accuracy. For CIFAR-100 and Tiny-ImageNet, NEO-KD saves $81.60\%$ and $50.27\%$ of computing budgets compared to \textit{Adv. w/o Distill}.
%In other words, AEOKD can attain high adversarial test accuracy with efficient budget usage.
%In Table \ref{table: Budget Saved}, we can see how much computing resources can be saved by AEOKD to output the most accurate predictions.
%For example, in MNIST dataset, AEOKD needs only $38.24\%$ budget ($4.29M$FLOPS) among total budget to achieve $90\%$ of adversarial test accuracy against max-average attack, compared that MaxAvg needs $56.12\%$ budget ($4.29M$FLOPS) and SKD needs $60.72\%$ budget ($6.813M$FLOPS). 
%AKD and EOKD also achieve higher adversarial test accuracy than MaxAvg and SKD except for CIFAR10.
%The reason of the exception in CIFAR10 dataset is that MaxAvg has higher adversarial test accuracy at exit 3 and 4 than SKD, AKD, and EOKD.
%Then, as more data are allocated at exit 3 and 4 (more budget is used), the performance of MaxAvg also increase.
%In fact, MaxAvg has lower adversarial test accuracy with low budget.
%Moreover, test accuracy of each method and adversarial test accuracy of each method built upon average attack are described in Supplementary Material.
The overall results are consistent with the results in anytime prediction setup,   confirming the advantage of our solution in practical settings  with limited computing budget. %Additional budgeted prediction experiments on MNIST dataset are provided in Supplementary Material.

\textbf{Result 3: Adversarial transferability.}\label{section: Transferability}
\begin{figure*}[!t]
	\centering
	\captionsetup[subfigure]{justification=centering}
	\begin{subfigure}{.245\textwidth}
		\includegraphics[width=1\linewidth]{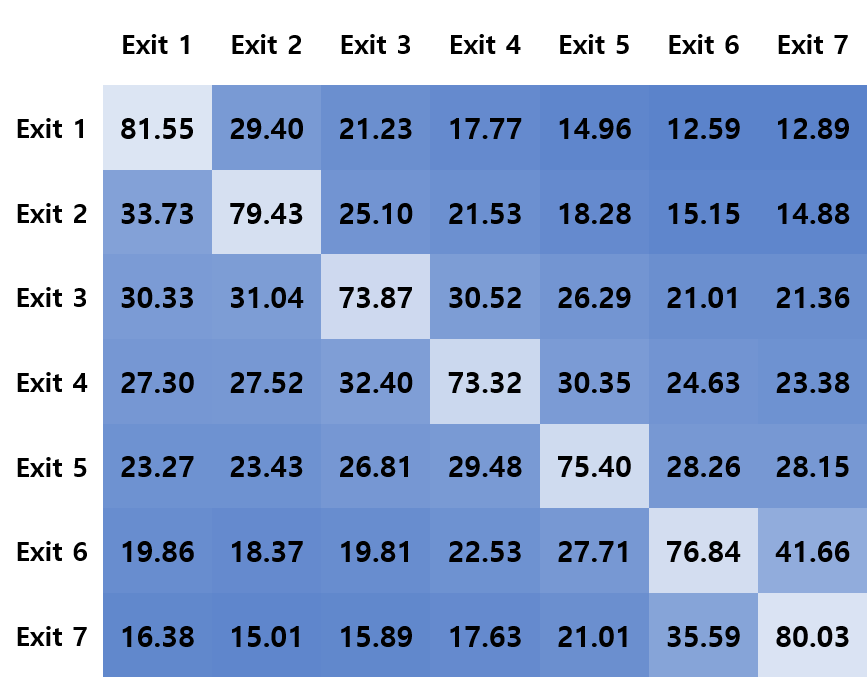}
		\caption{Adv. w/o Distill \cite{hu2019triple}\\ (Avg. w/o Diag.: 23.68\%)}
		\label{figure: MaxAvg AdvTrans}
	\end{subfigure}
	\begin{subfigure}{.245\textwidth}
		\includegraphics[width=1\linewidth]{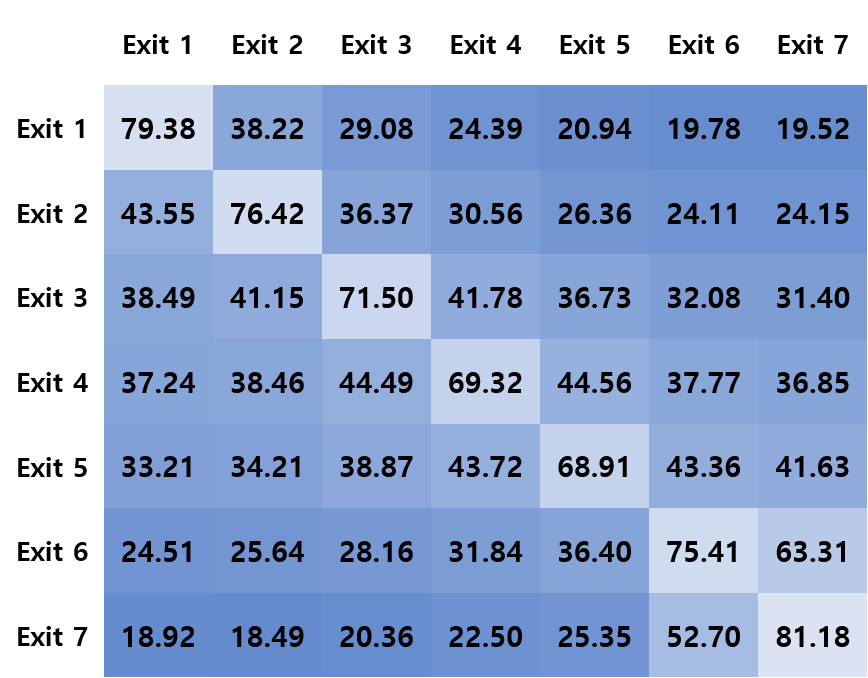}
		\caption{SKD \cite{phuong2019distillation} \\ (Avg. w/o Diag.: 33.36\%)}
		\label{figure: SKD AdvTrans}
	\end{subfigure}
	\begin{subfigure}{.245\textwidth}
		\includegraphics[width=1\linewidth]{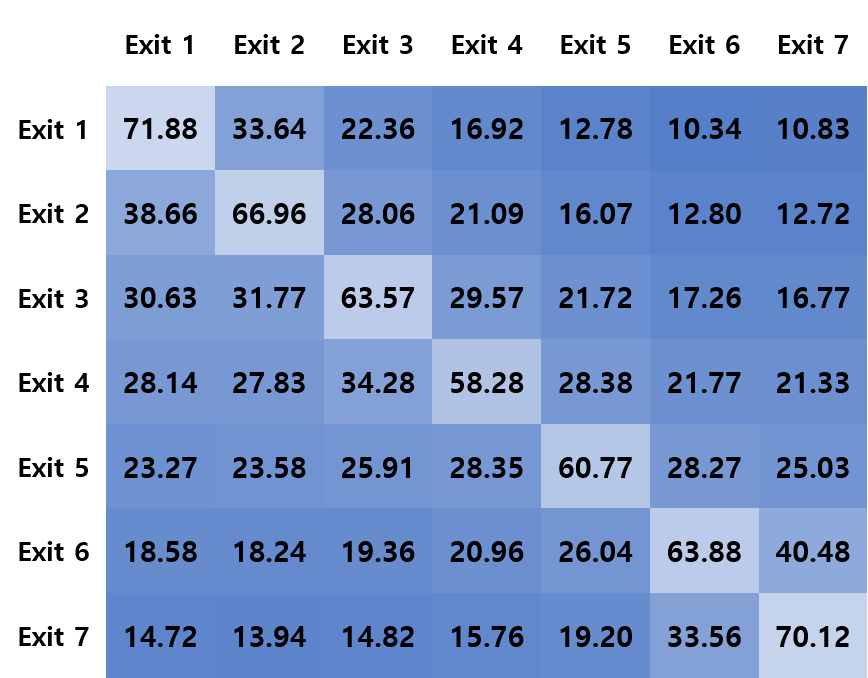}
		\caption{NKD \\ (Avg. w/o Diag.: 22.76\%)}
		\label{figure: NKD AdvTrans}
	\end{subfigure}
	\begin{subfigure}{.245\textwidth}
		\includegraphics[width=1\linewidth]{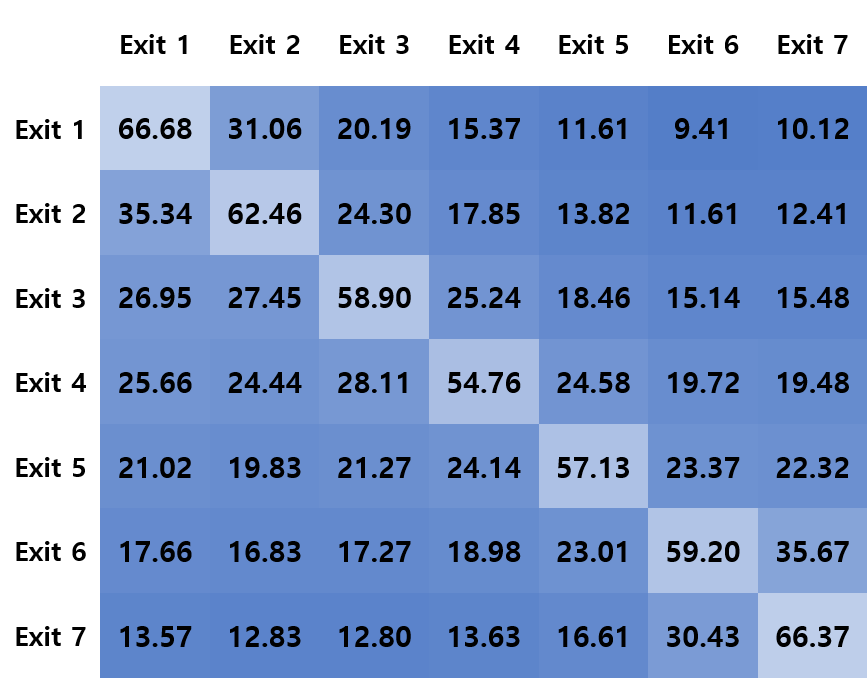}
		\caption{NEO-KD (ours) \\ (Avg. w/o Diag.: 20.12\%)}
		\label{figure: NEOKD AdvTrans}
	\end{subfigure}
 	\caption{\small \textbf{Adversarial transferability map} of each method on CIFAR-100.  Diag. indicates the diagonal of the matrix. \textbf{Row}: Target exit for generating adversarial examples. \textbf{Column}:  Exit where adversarial transferability is measured. Adopting NKD solely already achieves better adversarial transferability compared to the existing baselines. Applying EOKD to NKD can further improve adversarial transferability by reducing the dependency among different submodels in the multi-exit network. }
	\label{figure: AdvTrans}

\end{figure*}
We also compare the adversarial transferability of our NEO-KD and different baselines among exits in a multi-exit neural network. When measuring adversarial transferability,  as in \cite{yang2020dverge}, we initially gather all clean test samples for which all exits produce correct predictions. Subsequently, we generate adversarial examples targeting each exit using the collected clean samples (We use PGD-$50$ based single attack). Finally, we assess the adversarial transferability as the attack success rate of these adversarial examples at each exit.
%we first collect all the clean test samples that all exits give correct predictions.  Next, we generate adversarial examples targeting each exit with all collected clean samples.Lastly, we measure the adversarial transferability as the attack success rate of these adversarial examples at each exit. 
Figure \ref{figure: AdvTrans} shows the adversarial transferability map of each scheme on CIFAR-100. Here, each row corresponds to the target exit for generating adversarial examples, and each column corresponds the exit where attack success rate is measured. For example, the $(i,j)$-th element in the map is adversarial transferability measured at exit $j$, generated by the adversarial examples targeting exit $i$. The values and the brightness in the map indicate success rates of  attacks; lower value (dark color) means lower adversarial transferability. 
%\textcolor{red}{In addition, for convenient comparison, we notate the average of attack success rate values except for diagonal since attack success rate of diagonal is not fit to the definition of adversarial transferability.}

We have the following key observations from  adversarial transferability map. 
%First, from  Figure \ref{figure: MSDNet AdvTrans}, it can be seen that the multi-exit network without any adversarial training has high adversarial transferability, which makes the model to be vulnerable to single attack targeting a specific exit.
First, as observed in Figure \ref{figure: MaxAvg AdvTrans}, compared to \textit{Adv. w/o Distill} \cite{hu2019triple}, SKD \cite{phuong2019distillation} in Figure \ref{figure: SKD AdvTrans} exhibits higher adversarial transferability. This indicates that distilling the same teacher prediction to every exit leads to a high dependency across exits. Thus, it is essential to consider distilling non-overlapping knowledge across different exits. Second, when compared to the baselines \cite{hu2019triple, phuong2019distillation}, our proposed NKD in Figure \ref{figure: NKD AdvTrans} demonstrates low adversarial transferability. This can be attributed to the fact that NKD takes into account the quality of the distilled features and ensures that the features are not overlapping among exits. Third, as seen in Figure \ref{figure: NEOKD AdvTrans}, the adversarial transferability is further mitigated by incorporating  EOKD, which distills orthogonal class predictions to different exits, into  NKD. Comparing the average of attack success rate across all exits (excluding the values of the target exits shown in the diagonal), it becomes evident that NEO-KD yields $3.56\%$ and $13.24\%$ gains compared to \textit{Adv. w/o Distill} and SKD, respectively.
%First, compared to adversarial training without distillation  \cite{hu2019triple}  in Figure \ref{figure: MaxAvg AdvTrans}, SKD \cite{phuong2019distillation} in Figure \ref{figure: SKD AdvTrans} shows higher adversarial transferability. It means that distilling same teacher prediction to every exit results in high dependency across exits. Then, distilling the knowledge not to be overlapped across different exits should be considered.
%Second, compared to our baselines \cite{hu2019triple, phuong2019distillation}, our proposed NKD in Figure \ref{figure: NKD AdvTrans} shows low adversarial transferability. This is because NKD  considers the quality of the distilled features and distills the features not to be overlapped among exits. %the correlations among submodels simultaneously. 
%Third, as can be observed from Figure \ref{figure: NEOKD AdvTrans}, the adversarial transferability can be further improved by adopting our EOKD which distills orthogonal class predictions to different exits   to our NKD.  When comparing the attack success rate averaged over   all exits (except the values of the target exits shown in the diagonal), it can be seen that the NEO-KD has  3.68\% and 1.88\%   gains compared to \textit{Adv. w/o Distill} and SKD, respectively.
The overall results confirm the advantage of our solution to reduce the adversarial transferability in multi-exit networks. These results support the improved adversarial test accuracy of NEO-KD reported in Section \ref{sec:main_exp}. %5As a result, Figure \ref{figure: AdvTrans} shows an inverse proportion relationship between adversarial test accuracy and adversarial transferability. 
\subsection{Ablation Studies and Discussions}\label{subsec:ablation}
%\vspace{-1mm}
%by utilizing AKD and EOKD concurrently, the proposed AOEKD achieves comparable clean accuracy with
\textbf{Effect of each component of NEO-KD.}
\begin{table*}[t]
	\centering
	\begin{minipage}{.49\linewidth}
		\tiny
		\centering
		\begin{tabular}{c|ccc||c}
			\hline
			Exit & 1 & 2 & 3 & Average\\
			\hline
			Adv.   w/o Distill & 38.00\% & 41.42\% & 40.70\% & 40.04\% \\
			NKD & 40.01\% & 42.67\% & 40.23\% & 40.97\% \\
			EOKD & 37.26\% & 41.10\% & 38.68\% & 39.01\% \\
			\hline
			NEO-KD (ours) & \textbf{41.67\%} & \textbf{45.38\%} & \textbf{45.54\%} & \textbf{44.20\%} \\
			\hline
		\end{tabular}
		\captionsetup{justification=centering}
		\caption{\small \textbf{Effect of each component of NEO-KD}: Adversarial test accuracy against average attack on CIFAR-10. NKD and EOKD work in a complementary fashion and have implicit synergies. }
		\label{table: ablation CIFAR10}
	\end{minipage}%
	\centering
	\begin{minipage}{.49\linewidth}
		\tiny
		\centering
		\begin{tabular}{c|ccc||c}
			\hline
			Type of ensembles & 1 & 2 & 3 & Average\\
			\hline
			Adv.   w/o Distill & 38.00\% & 41.42\% & 40.70\% & 40.04\% \\
			\hline
			No ensembling & 39.54\% & 41.77\% & 40.15\% & 40.49\% \\
			Ensemble neighbors (NKD)& \textbf{41.67\%} &\textbf{45.38\%} & \textbf{45.54\%} & \textbf{44.20\%} \\
			Ensemble all exits & 38.63\% & 41.61\% & 39.93\% & 40.06\% \\
			\hline
		\end{tabular}
		\captionsetup{justification=centering}
		\caption{\small \textbf{Results with different number of ensembles}: Adversarial test accuracy against average attack on CIFAR-10. The neighbor-wise ensemble case corresponds to our NKD.}
		\label{table: number of group}
	\end{minipage} 
	\vspace{-3mm}
\end{table*}
\begin{table*}[t]
	\scriptsize
	\centering
	%\begin{subtable}[!t]{1\linewidth}
	\centering
	\begin{tabular}{c|c|ccccccc||c}
		\hline
		Attacks & Exit   & 1 & 2 & 3 & 4 & 5 & 6 & 7 & Average \\ \hline
		\multirow{2}{*}{PGD-100} & Adv. w/o Distill \cite{hu2019triple} &  16.82\% & 17.24\%  & 19.03\%  & 19.41\%  & 18.97\%  & 17.90\%  & 16.96\%  & 18.05\%    \\
		& NEO-KD (ours) & \textbf{20.38\%}  & \textbf{21.29\%}  & \textbf{23.22\%}  &  \textbf{24.38\%} & \textbf{24.44\%}  & \textbf{23.82\%}  & \textbf{22.21\%}  &  \textbf{22.82\%}   \\ \hline
		\multirow{2}{*}{CW} & Adv. w/o Distill \cite{hu2019triple} &  \textbf{35.31\%} & \textbf{35.72\%}  & 36.47\%  & 36.85\%  & 37.20\%  & 37.19\%  & 36.68\%  & 36.49\%    \\
		& NEO-KD (ours) & 31.56\%  & 34.55\%  & \textbf{38.20\%}  &  \textbf{40.60\%} & \textbf{43.48\%}  & \textbf{44.03\%}  & \textbf{43.01\%}  &  \textbf{39.35\%}   \\ \hline
		%\multirow{2}{*}{DeepFool} & Adv. w/o Distill &  \% & \%  &  \% & \%  & \%  & \%  & \%  & \%    \\
		%& NEO-KD (ours) & \textbf{\%}  & \textbf{\%}  & \textbf{\%}  &  \textbf{\%} & \textbf{\%}  & \textbf{\%}  & \textbf{\%}  &  \textbf{\%}   \\ \hline
		\multirow{2}{*}{AutoAttack} & Adv. w/o Distill \cite{hu2019triple} &  31.32\% & 34.24\%  & 37.27\%  & 39.67\%  & 41.20\%  & 41.73\%  & 40.45\%  & 37.98\%    \\
		& NEO-KD (ours) & \textbf{31.56\%}  & \textbf{34.55\%}  & \textbf{38.20\%}  &  \textbf{40.60\%} & \textbf{43.48\%}  & \textbf{44.03\%}  & \textbf{43.01\%}  &  \textbf{39.35\%}   \\ \hline
	\end{tabular}
	\hfill
	\newline
	\vspace{-1mm}
	\caption{\small \textbf{Results with stronger attacker algorithms}: Adversarial test accuracy against average attack using CIFAR-100 dataset.}
	\label{table: Stronger Attack}
	\vspace{-4mm}
\end{table*}
In Table  \ref{table: ablation CIFAR10}, we observe the effects of our individual components, NKD and EOKD. It shows that combining NKD and EOKD boosts up the performance beyond the sum of their original gains. Given different roles, the combination of NKD and EOKD enables multi-exit  networks to achieve the state-of-the-art performance under    adversarial attacks.

\textbf{Effect of the type of ensembles in NKD.} \label{section: group number}
In the proposed NKD, we consider only the neighbor exits to distill the knowledge of clean data.  What if we consider fewer or more exits than neighboring exits?  If the number of ensembles is too small, the scheme does not distill high-quality features. If the number of ensembles is too large, the dependencies among submodels increase, resulting in high adversarial transferability. To see this effect, in Table  \ref{table: number of group},  we measure adversarial test accuracy of  three types of ensembling methods depending on the number of exits used for constructing ensembles: \textit{no ensembling}, \textit{ensemble neighbors (NKD)}, and \textit{ensemble all exits}. In \textit{no enesmbling} approach, we distill the knowledge of each exit from clean data to the output at the same position of exit for adversarial examples. In contrast, the \textit{ensemble all exits} scheme averages the knowledge of all exits from clean data and provides it to all exits of adversarial examples.  The \textit{ensemble neighbors} approach corresponds to our NKD. 
The results show that the proposed NEO-KD with neighbor ensembling enables to distill high-quality features while lowering dependencies among submodels, confirming our intuition.

\textbf{Robustness against stronger adversarial attack.}
We evaluate  NEO-KD against stronger adversarial attacks; we perform average attack based on PGD-100 \cite{madry2018towards}, Carlini and Wagner (CW) \cite{carlini2017towards}, and AutoAttack \cite{croce2020reliable}. Table \ref{table: Stronger Attack} shows that  NEO-KD achieves higher adversarial test accuracy than \textit{Adv. w/o Distill} \cite{hu2019triple} in most of cases. Typically, CW attack and AutoAttack are  stronger attacks than the PGD attack in single-exit networks. However, in the context of multi-exit networks, these attacks become weaker than the PGD attack when taking  all exits into account. Details for generating stronger adversarial attacks are described in Appendix.%; given different roles, EOKD enables NKD to  \textcolor{red}{describe the synergy with insights!}
%\textcolor{red}{ 
%In this comparison, we can see that combining NKD and EOKD achieves higher adversarial test accuracy than adopting NKD or EOKD solely.  This phenomenon proves that there is an implicit synergy from combining NKD and EOKD.
%by balancing distillation losses. 
%Therefore, distilling both high quality features from clean data to adversarial examples and orthogonality across exits are important to improve robustness of multi-exit neural network.
%Additional experiment results with MNIST and CIFAR100 dataset are provided in Supplementary Material.}

\textbf{Additional results.} Other results including clean test accuracy, results with  average attack based adversarial training, results with varying hyperparameters, and results with another baseline used in single-exit network, %, and  results with different attacker algorithms with FGSM \cite{goodfellow2014explaining}, BIM \cite{kurakin2018adversarial}, MIM \cite{dong2018boosting},  Jitter \cite{schwinn2021exploring} method 
are provided in Appendix.

\section{Conclusion} 
%\vspace{-2mm}
In this paper, we proposed a new knowledge distillation based adversarial training strategy for robust multi-exit networks.  Our solution, NEO-KD,  reduces adversarial transferability in the network while guiding the output of the adversarial examples to closely follow the ensemble outputs of the neighbor exits of the clean data, significantly improving the overall adversarial test accuracy.  
%Moreover, NEO-KD   is a plug-and-play method which can be used in conjunction with other    training strategies tailored to multi-exit networks. 
Extensive experimental results on both anytime and budgeted prediction setups using various datasets confirmed the effectiveness of our method,  compared to baselines relying on existing
  adversarial training or knowledge distillation techniques for multi-exit networks. 

%\textbf{Limitations.} Considering our distillation method, NEO-KD is currently not directly applicable to object detection. Extending  our NEO-KD to other complicated tasks is an interesting future direction.

\section*{Acknowledgement}
This work was supported by the National Research Foundation of Korea (NRF) grant funded by the Korea government (MSIT) (No. NRF-2019R1I1A2A02061135), by the Center for Applied Research in Artificial Intelligence (CARAI) grant funded by DAPA and ADD (UD230017TD), and by IITP funds from MSIT of Korea (No. 2020-0-00626).

%%%%%%%%%%%%%%%%%%%%%%%%%%%%%%%%%%%%%%%%%%%%%%%%%%%%%%%%%%%%

\nocite{*}
\bibliography{main_paper_camera_ready_final.bib}

\newpage
\appendix

\renewcommand{\thetable}{A\arabic{table}}
\setcounter{table}{0}

\section{Experiment Details}
We provide additional implementation details that were not described in the main manuscript.
\subsection{Model training}
We train a SmallCNN \cite{hu2019triple} for $150$ epochs with batch size $128$  on MNIST \cite{lecun1998gradient}, a MSDNet \cite{huang2017multi} for $150$ epochs with batch size $128$ on CIFAR-10/100 \cite{krizhevsky2009learning} and Tiny-ImageNet \cite{le2015tiny}.
For ImageNet \cite{russakovsky2015imagenet}, since it takes a long time to train from the beginning, we finetune the pretrained model with our NEO-KD loss function for $10$ epochs.
All experiments are implemented with two RTX3090 GPUs.
Other settings follow \cite{hu2019triple}\footnote{https://github.com/VITA-Group/triple-wins} for SmallCNN and \cite{huang2017multi}\footnote{https://github.com/kalviny/MSDNet-PyTorch} for MSDNet except for a channel, which is set to $8$ for our experiments on CIFAR-10/100.
For the optimizer, SGD is used with a momentum of 0.9 and a weight decay of $5\times10^{-4}$.
For the MNIST dataset, the initial learning rate is set to $0.01$ and the learning rate is decayed $10$-fold at $50$ epoch.
For the CIFAR-10/100 dataset, the initial learning rate is $0.1$, and is decayed $10$-fold at $75$-th epoch and $115$-th epoch.
For Tiny-ImageNet, the initial learning rate is set to $0.1$, and is decayed $10$-fold at $50$-th epoch and $100$-th epoch.
For ImageNet, the learning rate is
  constant with $0.001$.

\subsection{Adversarial training}
During adversarial training, we use max-average attack and average attack \cite{hu2019triple} for generating adversarial examples via PGD attacker algorithm \cite{madry2018towards} with $7$-steps, while the PGD attacker algorithm with $50$-step is used for measuring robustness against a stronger attack during test time. 
With PGD-$50$ attack, we measure adversarial test accuracy on $3$ random seeds and average the results.
For PGD attack, the perturbation degree $\epsilon = 0.3$ is used for MNIST, $\epsilon = 8/255$ is used for CIFAR-10/100, and $\epsilon = 2/255$ Tiny-ImageNet/ImageNet during both adversarial training and inference time.
The step size $\delta$ is set to $20/255$ for MNIST, $2/255$ for CIFAR-10/100, and $\frac{2}{3}\epsilon$ (0.0052) for Tiny-ImageNet/ImageNet. The number of iterations is  commonly $50$-steps.
Similarly, when measuring adversarial transferability, we also use PGD attack with $50$ steps, $2/255$ step size, and $8/255$ epsilon value to generate adversarial attacks.
With PGD attack, the attack success rate is utilized as the metric to measure  adversarial transferability. We perform each experiment on $5$ random seeds and average the results.

Additionally, the hyperparameter $\alpha$ for NKD is set to $3$, and $\beta$ for EOKD is set to $1$ across all experiments.
On the other hand, the exit-balancing parameter $\gamma$ is set to $[1,1,1]$ for MNIST and CIFAR-10, and $[1,1,1,1.5,1.5]$, $[1,1,1,1.5,1.5,1.5,1.5]$ for Tiny-ImageNet/ImageNet, CIFAR-100, respectively.

%\subsubsection{Anytime Prediction Setup}
%\textcolor{red}{In anytime prediction setup, an appropriate exit can be selected  at inference stage depending on the latency constraint. The anytime prediction setup is effectively used in the time-constraint environment such as an emergency brake system in an autonomous driving, which must output fast predictions at an emergency situation.
	%In this setup, we measure the pure performance of each exit.
	%All test samples, which are corrupted by max-average/average attack, are predicted at every exit and we compute the average performance at each exit.
	%In addition, to compare the performance across our baselines and proposed methods, we also compute the average performance from all exits.}

\subsection{How to determine confidence threshold in budgeted prediction setup}

We provide a detailed explanation about how to determine confidence threshold for each exit using validation set before the testing phase. First, in order to obtain confidence thresholds for various budget scenarios, we allocate the number of validation samples for each exit. For simplicity, consider a toy example with 3-exit network (i.e., $L=3$) and assume the number of validation set is 3000. Then, each exit can be assigned a different number of samples: for instance, (2000, 500, 500), (1000, 1000, 1000) and (500, 1000, 1500). As more samples are allocated to the early exits, a scenario with a smaller budget can be obtained, while allocating more data to the later exits can lead to a scenario with a larger budget. More specifically, to see how to obtain the confidence threshold for each exit, consider the low-budget case of (2000, 500, 500). The model first makes predictions on all 3000 samples at exit 1 and sorts the samples based on their confidence. Then, the 2000-th largest confidence value is set as the confidence threshold for the exit 1. Likewise, the model performs predictions on remaining 1000 samples at exit 2 and the 500-th largest confidence is determined as the threshold for exit 2. Following this process, all thresholds for each exit are determined. During the testing phase, we perform predictions on test samples based on the predefined thresholds for each exit, and calculate the total computational budget for the combination of (2000, 500, 500). In this way, we can obtain accuracy and computational budget for different combinations of data numbers (i.e., various budget scenarios). Figures 2 and 3 in the main manuscript show the results for 100 cases of different budget scenarios.

\section{Clean Test Accuracy}
\begin{table*}[!t]
	\centering
	\begin{subtable}[!t]{1\linewidth}
		%		\small
		\centering
		\begin{tabular}{c|ccc||c}
			\hline
			Exit & 1 & 2 & 3 & Average\\
			\hline
			Adv. w/o Distill \cite{hu2019triple} & 98.14\% & 99.36\% & 99.55\% & 99.02\% \\
			%			SKD \cite{phuong2019distillation} & 98.11\% &\textbf{ 99.43\%} & 99.48\% & 99.01\% \\
			%			ARD \cite{goldblum2020adversarially} & 98.12\% & 99.27\% & 99.48\% & 98.96\%\\
			%			LW \cite{chen2022towards} & 97.87\% & 99.22\% & \textbf{99.49\%} & 98.86\% \\
			\hline
			NEO-KD (ours) & 97.82\% & 99.29\% & 99.48\% & 98.86\% \\
			\hline
		\end{tabular}
		\caption{MNIST}
		\label{table: Test Accuracy MNIST}
	\end{subtable}
	\hfill
	\newline
	\begin{subtable}[!t]{1\linewidth}
		\footnotesize
		\setlength{\tabcolsep}{3pt}
		\renewcommand{\arraystretch}{1.2}
		\centering
		\begin{tabular}{c|ccccc|c||ccccc|c}
			\hline
			& \multicolumn{6}{c||}{top-1 accuracy (\%)} & \multicolumn{6}{c}{top-5 accuracy (\%)} \\ \hline
			Exit & 1 & 2 & 3 & 4 & 5 & Avg. & 1 & 2 & 3 & 4 & 5 & Avg.\\
			\hline
			Adv.   w/o Distill  \cite{hu2019triple} & 43.10 & 47.64 & {50.36} & {50.58} & {50.70} & 48.48 & 68.26 & 72.30 & {74.86} & {75.58} & {75.36} & 73.27 \\
			%			SKD \cite{phuong2019distillation} & \textbf{48.08} & \textbf{50.46} & \textbf{50.90} & \textbf{51.14} & \textbf{49.92} & \textbf{50.10} & \textbf{73.22} & \textbf{75.62} & \textbf{76.22} & \textbf{75.66} & \textbf{75.10} & \textbf{75.16} \\
			%			ARD \cite{goldblum2020adversarially} & 43.46 & 47.72 & 49.28  & 50.10 & 50.12 & 48.14 & 69.34 & 72.70 & 74.72 & 74.82 & 74.06 & 73.13 \\
			%			LW \cite{chen2022towards} & 38.66 & 40.36 & 40.34 & 39.62 & 40.88 & 39.97 & 64.86 & 66.30 & 66.14 & 65.86 & 67.20 & 66.07 \\
			\hline 
			NEO-KD (ours) & {45.56} & {48.32} & 49.88 & 50.56 & 50.52 & {48.97} & {70.56} & {73.12} & 74.52 & 74.22 & 74.98 & 73.48 \\
			\hline
		\end{tabular}
		\caption{Tiny-ImageNet}
		\label{table: Test Accuracy Tiny-ImageNet}
	\end{subtable}
	\caption{\textbf{Clean test accuracy in the anytime prediction setup:} NEO-KD's advantage in terms of adversarial test accuracy (as shown in the main mansuscript) can be achieved without largely compromising the clean test accuracy.}
	\label{table: Test Accuracy}
\end{table*}

Table \ref{table: Test Accuracy} shows the clean accuracy results using the model built upon adversarial training via max-average attack. We observe that  NEO-KD generally shows comparable clean test accuracy with \textit{Adv. w/o Distill} \cite{hu2019triple}, especially on the more complicated dataset Tiny-ImageNet \cite{le2015tiny} while achieving much better adversarial test accuracy as reported in the main manuscript.

\section{Adversarial Training via Average Attack}
%\begin{table*}[!t]
%	\centering
%	\begin{subtable}[!t]{1\linewidth}
	%		\centering
	%		\begin{tabular}{c|ccc||c}
		%			\hline
		%			Exit & 1 & 2 & 3 & Average\\
		%			\hline
		%			Adv. w/o Distill \cite{hu2019triple} & \% & \% & \% & \% \\
		%			SKD \cite{phuong2019distillation} & \% & \% & \% & \% \\
		%			ARD \cite{goldblum2020adversarially} & \% & \% & \% & \% \\
		%			LW \cite{chen2022towards} & \% & \% & \% & \% \\
		%			\hline
		%			NEO-KD (ours) & \% & \% & \% & \% \\
		%			\hline
		%		\end{tabular}
	%		\caption{Max-average attack}
	%		\label{table: Avg AdvT MaxAvg Attack CIFAR10}
	%	\end{subtable}
%	\hfill
%	\newline
%	\begin{subtable}[!t]{1\linewidth}
	%		\centering
	%		\begin{tabular}{c|ccc||c}
		%			\hline
		%			Exit & 1 & 2 & 3 & Average\\
		%			\hline
		%			Adv. w/o Distill \cite{hu2019triple} & \% & \% & \% & \% \\
		%			SKD \cite{phuong2019distillation} & \% & \% & \% & \% \\
		%			ARD \cite{goldblum2020adversarially} & \% & \% & \% & \% \\
		%			LW \cite{chen2022towards} & \% & \% & \% & \% \\
		%			\hline
		%			NEO-KD (ours) & \% & \% & \% & \% \\
		%			\hline
		%		\end{tabular}
	%		\caption{Average attack}
	%		\label{table: Avg AdvT Avg Attack CIFAR10}
	%	\end{subtable}
%	\caption{\textbf{Anytime prediction setup:} Adversarial test accuracy on CIFAR-10 dataset.}
%	\label{table: Avg AdvT CIFAR10}
%\end{table*}

\begin{table*}[!t]
	\centering
	\begin{subtable}[!t]{1\linewidth}
		\scriptsize
		\setlength{\tabcolsep}{2.5pt}
		\centering
		\begin{tabular}{c|ccccccc|c|| ccccccc|c}
			\hline
			& \multicolumn{8}{c||}{top-1 accuracy (\%)} & \multicolumn{8}{c}{top-5 accuracy (\%)} \\ \hline
			Exit & 1 & 2 & 3 & 4 & 5 & 6 & 7 & Avg. & 1 & 2 & 3 & 4 & 5 & 6 & 7 & Avg.\\
			\hline
			Adv.   w/o Distill  \cite{hu2019triple} & 21.14 & 20.49 & 20.68 & 20.18 & 20.58 & 20.56 & 20.24 & 20.55 & 48.28 & 48.24 & 46.75 & 46.41 & 46.31 & 46.96 & 45.64 & 46.94 \\
			SKD \cite{phuong2019distillation} & \textbf{22.39} & \textbf{22.17} & 19.89 & 18.81 & 18.82 & 17.48 & 14.84 & 19.20 &\textbf{51.73 }&\textbf{51.43} & 47.88 & 45.92 & 45.90 & 43.65 & 41.38 & 46.84 \\
			ARD \cite{goldblum2020adversarially} & 20.55  & 19.84 & 20.95 & 19.61 & 19.41 & 20.52 & \textbf{21.72} & 20.37 & 48.18& 46.94 & 46.96 & 45.37 & 45.06 & 46.17 & \textbf{48.42} & 46.73 \\
			LW \cite{chen2022towards} & 17.65  & 18.80 & 16.97  & 16.64 & 17.15 & 17.08 & 16.79 & 17.30 & 44.33 & 44.03 & 41.49 & 41.48 & 41.37 & 42.06 & 41.64 & 42.34 \\
			\hline 
			NEO-KD (ours) & 21.31  & 22.01 & \textbf{21.46} & \textbf{22.10} & \textbf{21.89} & \textbf{21.64} & 20.78 & \textbf{21.60} & 49.16 & 48.88 & \textbf{49.12} & \textbf{48.76} & \textbf{48.00} & \textbf{48.08}& 47.23 & \textbf{48.46} \\
			\hline
		\end{tabular}
		\caption{Max-average attack}
		\label{table: Avg AdvT MaxAvg Attack}
	\end{subtable}
	\begin{subtable}[!t]{1\linewidth}
		\scriptsize
		\setlength{\tabcolsep}{2.5pt}
		\centering
		\begin{tabular}{c|ccccccc|c|| ccccccc|c}
			\hline
			& \multicolumn{8}{c||}{top-1 accuracy (\%)} & \multicolumn{8}{c}{top-5 accuracy (\%)} \\ \hline
			Exit & 1 & 2 & 3 & 4 & 5 & 6 & 7 & Avg. & 1 & 2 & 3 & 4 & 5 & 6 & 7 & Avg.\\
			\hline
			Adv.   w/o Distill  \cite{hu2019triple} & 12.73 & 12.10 & 13.33 & 13.27 & 13.14 & 13.17 & 13.27 & 13.00 & 36.22 & 35.66 & 37.07 & 36.57 & 36.47 & 37.26 & 36.51 & 36.54\\
			SKD \cite{phuong2019distillation} & \textbf{15.96} & \textbf{ 15.90} & 15.31 & 15.32 & 15.70 & 14.84 & 10.33 & 14.77 &\textbf{42.19} & \textbf{41.72} & \textbf{41.29} & \textbf{41.46} & \textbf{42.03} & 40.08 &32.68 & \textbf{40.21} \\
			ARD \cite{goldblum2020adversarially} & 12.74  & 12.53 & 13.80 & 13.04 & 12.79& 13.56 & 14.73 & 13.31 & 36.12 & 35.67 & 37.21 & 36.28 & 35.81 & 37.09 & 38.82 & 36.71 \\
			LW \cite{chen2022towards} & 12.82  & 13.33 & 12.28  & 11.96 & 12.10 & 12.25 & 12.55 & 12.47 & 36.49 & 36.23 & 34.46 & 34.17 & 34.09 & 34.78 & 34.72 & 34.99 \\
			\hline 
			NEO-KD (ours) & 14.64  &15.15 & \textbf{15.42} & \textbf{16.07} &\textbf{ 16.42} & \textbf{16.11} & \textbf{15.59} & \textbf{15.63} & 38.88 & 39.56 & 40.60 & 40.57 & 41.13 & \textbf{40.91} & \textbf{39.85} & \textbf{40.21} \\
			\hline
		\end{tabular}
		\caption{Average attack}
		\label{table: Avg AdvT Avg Attack}
	\end{subtable}
	\caption{\textbf{Adversarial training via Average attack:} Adversarial test accuracy in the anytime prediction setup on CIFAR-100.}
	\label{table: Avg AdvT}
\end{table*}

In the main manuscript, we presented experimental results using the model trained based on max-average attack. Here, we also adversarially train the model via average attack \cite{hu2019triple} and measure adversarial test accuracy on  CIFAR-100 dataset.
Table \ref{table: Avg AdvT} compares adversarial test accuracies of NEO-KD and other baselines against max-average attack and average attack. The overall results are consistent with the ones in the main manuscript with adversarial training via max-average attack, further confirming the advantage of NEO-KD.

%These comparisons also show the similar role of our proposed method as adversarial training via max-average attack. 
%SKD, which utilizes existing self-distillation scheme, achieves lower adversarial test accuracy than other schemes in most of cases.
%For our schemes, both AKD and EOKD achieve higher adversarial test accuracy than Adv. w/o Distill and SKD in CIFAR-100, while EOKD has same weakness where each exit is distilled only one non-maximal class as adversarial training via max-average attack in CIFAR-10.
%In spite of this shortcoming in CIFAR-10, combined method (AEOKD) achieves the best adversarial test accuracy  by implicit synergy from balancing the distillation loss.
%Therefore, Table \ref{table: Avg AdvT} and \ref{table: Avg AdvT CIFAR10} prove that our AEOKD is effective not only  for adversarial training with max-average attack but also for adversarial training with average attack.

\section{Hyperparameter Tuning}
In the NEO-KD objective function, there are three hyperparameters ($\alpha, \beta, \gamma$), where
$\alpha$, $\beta$ control the amount of distilling knowledge from NKD, EOKD and $\gamma$ increases the amount of knowledge distilled to later exits.

\subsection{Hyperparameter $(\alpha, \beta)$}
%We choose the 
The extreme value of $\alpha$ and $\beta$ can destroy ideal adversarial training.
Too large $\alpha$ makes strong NKD, which results in high dependency among submodels and too small $\alpha$ makes weak NKD, which cannot distill enough knowledge to student exits.
In contrast, too large $\beta$ makes strong EOKD, which can interrupt adversarial training by distilling only sparse knowledge (likelihoods of majority classes are zero) and too small $\beta$ makes weak EOKD, which cannot mitigate dependency among submodels.
We select $\alpha$, $\beta$ values in the range of $[0.35, 3]$ and measure the adversarial test accuracy value by averaging adversarial test accuracy from all exits. 
The candidate $(\alpha, \beta)$ pairs are $(0.35, 1)$, $(1, 0.35)$, $(0.35, 0.35)$, $(0.5, 1)$, $(1, 0.5)$, $(0.5, 0.5)$, $(1, 1)$, $(2, 1)$, $(1, 2)$, $(2, 2)$, $(3, 1)$, $(1, 3)$, and $(3, 3)$.
When $(\alpha, \beta)$ is $(3, 1)$, NEO-KD achieves $28.96\%$ of adversarial test accuracy against max-average attack and $22.88\%$ against average attack, which is the highest adversarial test accuracy among the various candidate $(\alpha, \beta)$ pairs.
Therefore, we use $(3, 1)$ as $(\alpha, \beta)$ pair in our experiments.

\subsection{Hyperparameter $\gamma$}
Since the prediction difference between the last exit (teacher prediction) and later exits  is smaller than the prediction difference between the last exit and early exits, later exits are less effective for taking advantage of knowledge distillation.
Therefore, we provide slightly larger weights to later exits for distilling more knowledge to later exits than early exits.
The candidate $\gamma$ values are $[1, 1, 1, 1, 1, 1, 1]$, $[1, 1, 1, 1.5, 1.5, 1.5, 1.5]$, and $[1, 1, 1, 1.7, 1.7, 1.7, 1.7]$.
As a result, when we distill $1.5$ times more knowledge to later exits, NEO-KD achieves $28.96\%$ of adversarial test accuracy against max-average attack and $22.88\%$ against average attack, which is the highest adversarial test accuracy compared to providing same weights with earlier exits to later exits ($28.13\%$ for max-average and $21.66\%$ for average attack) or distilling $1.7$ times more knowledge to later exits than earlier exits ($28.68\%$ for max-average and $22.58\%$ for average attack).
The adversarial test accuracy value is the average of adversarial test accuracies from all exits.
Therefore, we use $\gamma = [1, 1, 1, 1.5, 1.5, 1.5, 1.5]$ in our experiments. This result proves that the exit-balancing parameter $\gamma$ with an appropriate value is needed for high performance.

\section{Discussions on Performance Degradation at Later Exits}
As can be seen from the results for the anytime prediction in the main manuscript, the adversarial test accuracy of the later exits is sometimes lower than the performance of earlier exits. This phenomenon can be explained as follows: In general, we observed via experiments that adversarial examples targeting later exits has the higher sum of losses from all exits compared to adversarial examples targeting earlier exits. This makes max-average or average attack mainly focus on attacking the later exits, leading to   low adversarial test accuracy at  later exits. The performance of later exits can be improved by adopting the ensemble strategy as in the main manuscript for the budgeted prediction setup.

\section{Comparison with Recent Defense Methods for Single-Exit Networks}

\begin{table*}[!t]
	\centering
	\begin{subtable}[!t]{1\linewidth}
		%		\small
		\centering
		\begin{tabular}{c|ccc||c}
			\hline
			Exit & 1 & 2 & 3 & Average\\
			\hline
			PGD-TE \cite{dong2021exploring} & \textbf{48.73\%} & 46.00\% & 46.85\% & 47.19\% \\
			TRADES-TE \cite{dong2021exploring} & 45.05\% & 39.64\% & 42.10\% & 42.26\% \\
			\hline
			NEO-KD (ours) & 46.53\% & \textbf{47.65\%} & \textbf{50.71\%} & \textbf{48.30\%} \\
			\hline
		\end{tabular}
		\caption{CIFAR-10}
		\label{table: TEAT CIFAR-10}
	\end{subtable}
	\hfill
	\newline
	\begin{subtable}[!t]{1\linewidth}
		\footnotesize
		\setlength{\tabcolsep}{3pt}
		\renewcommand{\arraystretch}{1.2}
		\centering
		\begin{tabular}{c|ccccccc||c}
			\hline
			Exit & 1 & 2 & 3 & 4 & 5 & 6 & 7 & Avg.\\
			\hline
			PGD-TE  \cite{dong2021exploring} & 24.07\% & 24.39\% & 25.14\% & 25.35\% & 26.29\% & 25.57\% & 24.60\% & 25.06\% \\
			TRADE-TE \cite{dong2021exploring} & 17.62\% & 18.52\% & 18.61\% & 18.98\% & 18.95\% & 19.67\% & 20.35\% & 18.96\% \\
			\hline 
			NEO-KD (ours) & \textbf{28.37\%} & \textbf{28.78\%} & \textbf{29.02\%} & \textbf{29.49\%} & \textbf{30.06\%} & \textbf{28.45\%} & \textbf{28.54\%} & \textbf{28.96\%} \\
			\hline
		\end{tabular}
		\caption{CIFAR-100}
		\label{table: TEAT CIFAR-100}
	\end{subtable}
	\caption{Comparison of adversarial test accuracy against max-average attack between TEAT methods and our NEO-KD.}
	\label{table: TEAT}
\end{table*}

The baselines in the main paper were generally the adversarial defense methods designed for multi-exit networks. In this section, we conduct  additional experiments with a recent defense method, TEAT \cite{dong2021exploring}, and compare with our method. Since TEAT was originally designed for the single-exit network, we first adapt TEAT to the multi-exit network setting. Instead of the original TEAT that generates the adversarial examples considering the final output of the network, we modify TEAT to generate adversarial examples that maximizes the average loss of all exits in the multi-exit network. Table \ref{table: TEAT} below shows the results using max-average attack on    CIFAR-10/100. It can be seen that our NEO-KD, which is designed for multi-exit networks, achieves higher adversarial test accuracy compared to the TEAT methods (PGD-TE and TRADES-TE) designed for single-exit networks. The results highlight the necessity of developing adversarial defense techniques geared to multi-exit networks rather than adapting general defense methods used for single-exit networks.

\section{Comparison with SKD and ARD}

\begin{table*}[!t]
	\centering
	\hfill
	\newline
	\begin{tabular}{c|ccc||c}
		\hline
		Exit & 1 & 2 & 3 & Avg.\\
		\hline
		SKD (exit 1) & 32.27\% & 36.92\% & 38.57\% & 35.92\% \\
		SKD (exit 2) & 35.33\% & 35.10\% & 37.82\% & 36.08\% \\
		SKD (exit 3) & 39.36\% & 41.39\% & 38.39\% & 39.71\% \\
		SKD (ensemble) & 38.63\% & 41.80\% & 40.13\% & 40.19\% \\
		\hline 
		ARD (exit 1) & 35.64\% & 38.10\% & 42.12\% & 38.62\% \\
		ARD (exit 2) & 35.35\% & 38.24\% & 40.00\% & 37.86\% \\
		ARD (exit 3) & 39.37\% & 41.98\% & 43.53\% & 41.63\% \\
		ARD (ensemble) & 35.22\% & 38.35\% & 40.76\% & 38.11\% \\
		\hline
		\hline
		NEO-KD (ours) & 41.67\% & 45.38\% & 45.54\% & 44.20\%\\
		\hline
	\end{tabular}
	\caption{\normalsize Adversarial test accuracy of SKD and ARD according to exit selection as a teacher prediction.}
	\label{table: Exit Select}
\end{table*}

Existing self-distillation schemes \cite{li2019improved,phuong2019distillation} for multi-exit networks improve the performance on clean samples by self-distilling the knowledge of the last exit, as the last exit has the best prediction quality. Therefore, following the original philosophy, we also used the last exit in implementing the SKD baseline. Regarding ARD \cite{goldblum2020adversarially}, since it was proposed for single-exit networks, we also utilized the last exit with high performance when applying ARD to multi-exit networks. Nevertheless, we perform additional experiments to consider comprehensive baselines using various exits for distillation. Table \ref{table: Exit Select} above shows the results of SKD and ARD using a specific exit or an ensemble of all exits for distillation. The results show that our scheme consistently outperforms all   baselines.

\section{Implementations of Stronger Attacker Algorithms}
In Section 4.3 of the main manuscript, during inference, we replaced the Projected Gradient Descent (PGD) attack with other attacker algorithms (PGD-$100$ attack, Carlini and Wagner (CW) attack \cite{carlini2017towards}, and AutoAttack \cite{croce2020reliable}) to generate stronger attacks for multi-exit neural networks. This section provides explanation on how these stronger attacks are implemented tailored to multi-exit neural networks.

\subsection{Carlini and Wagner (CW) attack}
The Carlini and Wagner (CW) attack is a method of generating adversarial examples designed to reduce the difference between the logits of the correct label and the largest logits among incorrect labels. In alignment with this attack strategy, we modify the CW attack for multi-exit neural networks. In the process of minimizing this difference, our modification aims to minimize the average difference across all exits of the multi-exit neural network. Moreover, when deciding whether a sample has been successfully converted into an adversarial example, we consider a sample adversarial if it misleads all exits in the multi-exit neural network.

\subsection{AutoAttack}
AutoAttack produces adversarial attacks by ensembling various attacker algorithms. For our experiment, we sequentially use APGD \cite{croce2020reliable}, APGD-T \cite{croce2020reliable}, FAB \cite{croce2020minimally}, and Square \cite{andriushchenko2020square} algorithms to generate adversarial attacks, as they are commonly used.

\subsubsection{APGD and APGD-T}
%APGD attack is a modified version of PGD attack. PGD attack has limitations in that fixed step size, which is a suboptimal selection. APGD resolves the limitations by adopting adaptive step size and momentum term. Similarly, APGD-T attack is a modified version of APGD attack, where the attack perturbs a sample to be changed a specific class. In this process, we use the average loss of all exits in multi-exit neural network as the loss for computing gradient for adversarial example update. In addition, when deciding whether a sample is successfully converted into an adversarial example, we determine that a sample is adversarial if it successfully misleads all exits in the multi-exit neural network.
The APGD attack is a modified version of the PGD attack, which is limited by its fixed step size, a suboptimal choice. The APGD attack overcomes this limitation by introducing an adaptive step size and a momentum term. Similarly, the APGD-T attack is a variation of the APGD attack where the attack perturbs a sample to change to a specific class. In this process, we use the average loss of all exits in the multi-exit neural network as the loss for computing the gradient for adversarial example updates. Moreover, we define a sample as adversarial if it misleads all exits in the multi-exit neural network.

\subsubsection{FAB}
%FAB attack generates adversarial attacks by the linear approximation of classifiers, projection to the classifier hyperplane, convex combinations, and extrapolation. During the generation process, FAB attack defines a hyperplane classifier dividing two classes and finds a new adversarial example by convex combination of extrapolation-projected current adversarial example and extrapolation-projected original sample with minimum perturbation norm. In the process, we use the average gradient of all exits in multi-exit neural network as the gradient for updating adversarial examples. In addition, when deciding whether a sample is successfully converted into an adversarial example, we determine that a sample is adversarial if it successfully misleads all exits in the multi-exit neural network.
The FAB attack creates adversarial attacks through a process involving linear approximation of classifiers, projection to the classifier hyperplane, convex combinations, and extrapolation. The FAB attack first defines a hyperplane classifier separating two classes, then finds a new adversarial example through a convex combination of the extrapolation-projected current adversarial example and the extrapolation-projected original sample with the minimum perturbation norm. Here, we use the average gradient of all exits in the multi-exit neural network as the gradient for updating adversarial examples. Similar to above, we label a sample as adversarial if it can mislead all exits in the multi-exit neural network.

\subsubsection{Square}
%Square attack generates adversarial attacks by random searches of adversarial patches with random perturbation degree and random position. Square attack algorithm iteratively sample the perturbation degree and the positions of patches randomly while decreasing the size of patches. The sampled adversarial patches are injected to a sample and selected one that increases the loss of the target model the most. In the process, we use the average loss of all exits in multi-exit neural network as the loss for identifying that a sampled perturbation increases the loss of the target model or decreases the loss of target model to target class. In addition, when deciding whether a sample is successfully converted into an adversarial example, we determine that a sample is adversarial if it successfully misleads all exits in the multi-exit neural network.
The Square attack generates adversarial attacks via random searches of adversarial patches with variable degrees of perturbation and position. The Square attack algorithm iteratively samples the perturbation degree and the positions of patches while reducing the size of patches. The sampled adversarial patches are added to a sample, and the  patches that maximizes the loss of the target model are selected. Here, we use the average loss of all exits in the multi-exit neural network as the loss for determining whether a sampled perturbation increases or decreases the loss of the target model. Additionally, we determine a sample as adversarial if it misleads all exits in the multi-exit neural network.
\subsection{Experiment details}
%We commonly use $\epsilon = 0.03$ as a perturbation degree and generating adversarial examples for $50$ steps. All the attacks are L-$\infty$ norm based attacks. In APGD attack, we use cross entropy loss for computing the gradient for updating adversarial examples. In APGD-T attack and FAB attack, we consider all class pairs for generating adversarial attacks. In Square attack, random search operation is conducted for $5000$ times (the number of queries). 
%For the performance comparison against stronger attacker algorithms, adversarial training via average attack is adopted in Adv. w/o Distill \cite{hu2019triple} and NEO-KD (ours). However, since the original CW attack algorithm and AutoAttack algorithm were designed for single-exit neural networks, this adapted versions targeting multi-exit neural networks are relatively weaker.
For all attacks, we commonly use $\epsilon = 0.03$ as the perturbation degree and generate adversarial examples over $50$ steps. All the attacks are based on $L_{\infty}$ norm. For the APGD attack, we employ cross entropy loss for computing the gradient to update adversarial examples. In both the APGD-T and FAB attacks, all class pairs are considered when generating adversarial attacks. For the Square attack, the random search operation is conducted 5000 times (the number of queries). Other settings follow \cite{kim2020torchattacks}. In terms of performance comparison against stronger attacker algorithms, we adopt adversarial training via average attack in both Adv. w/o Distill \cite{hu2019triple} and NEO-KD (our approach). However, since the original CW attack algorithm and AutoAttack algorithm were designed for single-exit neural networks, this adapted versions targeting multi-exit neural networks are relatively weak.

%Not only from adversarial attack generated by PGD method, our AEOKD shows high robustness against FGSM, BIM, MIM, Jitter methods. Also, our AEOKD achieves higher adversarial test accuracy than adversarial training without knowledge distillation in most of cases. With these results against various attacker algorithms, experiment results prove that our AEOKD is robust against another attack as well as PGD attack and generally more robust than adversarial training without knowledge distillation.

\end{document}